\documentclass[10pt,journal,compsoc]{IEEEtran}
\ifCLASSOPTIONcompsoc
  \usepackage[nocompress]{cite}
\else
  \usepackage{cite}
\fi
\ifCLASSINFOpdf
  \usepackage[pdftex]{graphicx}
  \DeclareGraphicsExtensions{.pdf,.jpeg,.png}
\else
\fi
\usepackage{amsmath}
\usepackage{amssymb}

\usepackage{footnote}
\makesavenoteenv{tabular}

\usepackage[export]{adjustbox}

\usepackage{booktabs}
\usepackage{bm}
\usepackage[dvipsnames]{xcolor}
\usepackage{lscape}
\usepackage{bbding}
\usepackage{authblk}

\usepackage{multirow}
\usepackage{array}
\usepackage{placeins}

\usepackage[normalem]{ulem}
\usepackage{threeparttable}

\usepackage{hyperref}
\hypersetup{
    colorlinks=false,
}
\renewcommand{\v}[1]{\mathbf{#1}}
\newcommand{\vg}[1]{\bm{#1}}

\newcommand{\xstar}{\v{x}^\star}
\newcommand{\pool}{\mathsf{pool}}
\newcolumntype{C}[1]{>{\centering\arraybackslash}p{#1}}
\newcolumntype{R}[1]{>{\raggedleft\arraybackslash}p{#1}}
\newcolumntype{L}[1]{>{\raggedright\arraybackslash}p{#1}}

\newcommand{\clineh}[1]{\cline{#1}\rule{0pt}{3ex}}

\newcommand{\remove}[1]{}

\definecolor{revcolor}{RGB}{25,25,200}

\begin{document}
%
\title{Fine-grained Species Recognition with Privileged Pooling: Better Sample Efficiency Through Supervised Attention}

\author[1]{Andr\'es C. Rodr\'iguez}
\author[1]{Stefano D'Aronco}
\author[1]{Konrad Schindler}
\author[1,2]{Jan Dirk Wegner}
\affil[1]{EcoVision Lab - Photogrammetry and Remote Sensing, ETH Zurich, Switzerland}
\affil[2]{Institute for Computational Science, University of Zurich, Switzerland}

\IEEEtitleabstractindextext{%
\begin{abstract}
We propose a scheme for supervised image classification that uses privileged information, in the form of keypoint annotations for the training data, to learn strong models from small and/or biased training sets. 
Our main motivation is the recognition of animal species for ecological applications such as biodiversity modelling, which is challenging because of long-tailed species distributions due to rare species, and strong dataset biases such as repetitive scene background in camera traps.
To counteract these challenges, we propose a visual attention mechanism that is supervised via keypoint annotations that highlight important object parts.
This privileged information, implemented as a novel privileged pooling operation, is only required during training and helps the model to focus on regions that are discriminative.
In experiments with three different animal species datasets, we show that deep networks with privileged pooling can use small training sets more efficiently and generalize better.
\end{abstract}

}

\maketitle

\IEEEdisplaynontitleabstractindextext

%
\IEEEpeerreviewmaketitle

\section{Introduction}

Learning under privileged information is a paradigm where, exclusively for the training samples, one has access to supplementary information beyond the target outputs~\cite{lupi_2010,yang2017miml,lee2018spigan,lambert2018deep}. The idea is to use this side information to guide the training towards a model that achieves lower generalization error. Such an approach can be beneficial in two situations: \emph{(i)} compared to standard supervised learning it is in general possible to achieve better performance with the same (typically small) number of training samples; \emph{(ii)} it is possible to steer the learning so as to overcome potential biases in the training set. {Both situations arise in many domains, but are particularly challenging for the fine-grained classification of animal species: due to difficulties of observing and photographing animals, practical training sets will unavoidably suffer from observational biases and also have limited sample sizes for certain classes.}

The concept of privileged information during training was originally introduced in~\cite{lupi_2010} to improve the estimation of slack variables and the convergence rate of Support Vector Machines (SVMs). Subsequent works~\cite{yang2017miml,lee2018spigan,lambert2018deep} have adapted this  idea to a variety of visual tasks, by adding bounding boxes, attributes or sketches as privileged information \cite{IB_visual2016,chen2017training}. Technically, one can interpret {learning under} privileged information as a regularization of the model parameters with additional knowledge about the training samples.

Many common CNN architectures, like ResNet~\cite{he2016} or Inception~\cite{szegedy2015}, employ a global average pooling layer before the final fully connected layer(s), in order to reduce the number of parameters and to make the model applicable to input images of varying size. However, much information is lost during feature averaging, as features of the object of interest (in our case the animal) are merged with background features. Evidently, this can lead to noisy representations, particularly if the training set for some target classes is small -- a 
frequent situation when dealing with skewed data distributions where some classes, for instance certain animal species~\cite{Horn_2018_CVPR}, are much rarer than others.
A similar problem arises when the procedure used to acquire the training data induces sampling biases, which then may cause the network to learn spurious correlations that are irrelevant, or even harmful, for the task~\cite{torralba2011unbiased,ritter2017cognitive}.
Specifically, global pooling operations harm generalization if different categories of interest often appear in the same context, thus complicating the conceptually simple task to focus on a small, relevant region; such as for instance in our application, where animals are surrounded by similar vegetation. 
We thus advocate the use of privileged information during training to guide the model's attention.

\begin{figure}[t!]
\centering
\includegraphics[width=\columnwidth, trim=5cm 5.5cm 0cm 0cm, clip]{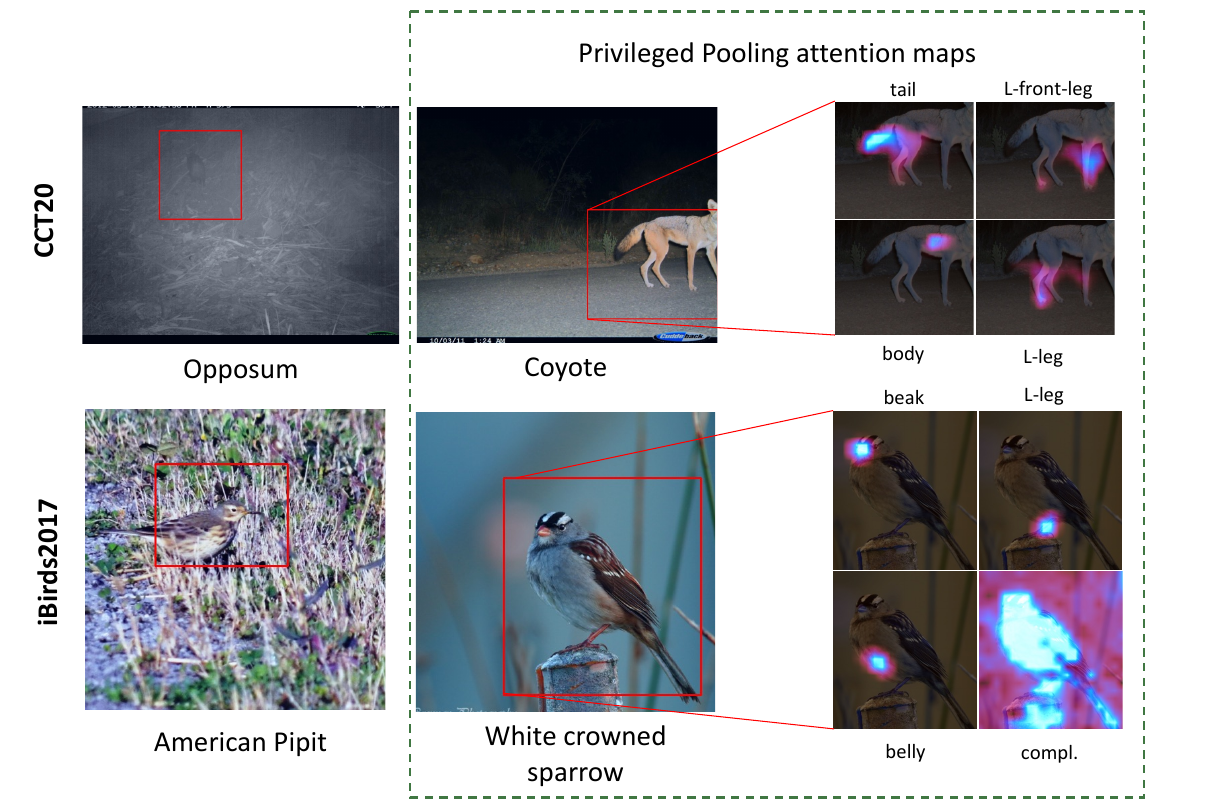}
\caption{Predicted classes using privileged pooling on CCT20-Cis test dataset (top) and iBirds2017 test dataset (bottom). Bounding boxes are computed using the predicted attention maps. Attention maps (bounding-box cropped for visualization) depict the encoded privileged information from different keypoints provided at train time. The bottom right-most attention map is not supervised by any keypoint and acts as complementary to other animal regions.
}
\label{fig:teaser}
\end{figure}

We introduce \emph{privileged pooling (PrPool)} a visual attention mechanism {for animal species recognition} that leverages privileged information in the form of keypoint locations to learn a weighted pooling operator.
It is intuitive that annotations of important {body parts} facilitate learning from small training sets. We use point-level part annotations, which are relatively cheap to collect and at the same time directly relevant to discriminate {animal species} that look alike. 
A few works have investigated self-supervised attention as a means to improve image understanding~\cite{hu2019see,zheng2017learning,tellmewheretolook_2018}. We are not aware of any systematic inquiry into supervised learning of the pooling operator from privileged information.
Notably,~\cite{attentional_pooling_girdhar_NIPS2017} derive a scheme for action recognition that relies on keypoints to learn a very specific pooling operation (a low-rank approximation of bilinear pooling).

Here, we explore the role of attention maps as a general tool to capture spatially explicit privileged information. {In our case, the additional annotations come in the form of {animal} keypoints, which are used to train attention maps as a soft gating mechanism that selects relevant features.}
{Fig.~\ref{fig:teaser} shows an example how such an attention map emphasizes different keypoints on a test sample, like the tail, head, and body.}
In this way we obtain a generic attention module that can be combined with any commonly used pooling operator to improve classification performance. 
{We also provide two new datasets with part annotations: (1) Caltech CameraTrap-20+ (CCT20+), obtained by augmenting a subset of the Caltech CameraTrap-20 dataset~\cite{Beery_2018_ECCV}. With this challenging dataset, we demonstrate that our method is able to counteract inherent background biases and thereby improve classification performance. (2) iBirds2018+, a subset of rare bird species  from iNaturalist2018~\cite{Horn_2018_CVPR}. With this dataset we show that our method improves classification in the long tail of rarely observed species.}
Furthermore, we also experiment with the CUB200~\cite{WelinderCUB2010} birds dataset under a scarce data regime, and also outperform {state-of-the-art methods} based on both privileged information and few-shot learning.
To assess generalization, we extract the matching  subset of the \emph{aves} (bird) family {(termed \textit{iBirds2017})} from the iNaturalist-17 dataset~\cite{Horn_2018_CVPR} and test our model trained on CUB200.
In that experiment the advantage of our model is even bigger.
Overall, we show that supervising the pooling with privileged information affords better generalization with fewer training samples, and is also a powerful alternative to few-shot learning when labeled training data is scarce.
{Code and the corresponding CCT20+ {and iBirds2018+} annotations will be made available at \href{https://github.com/ac-rodriguez/privilegedpooling}{github.com/ac-rodriguez/privilegedpooling}.}

\section{Related work}\label{sec:related_work}

\begin{figure}[t]
    \centering
   \includegraphics[width=\columnwidth, trim={0.5cm 2cm 1.9cm 0.5cm},clip]{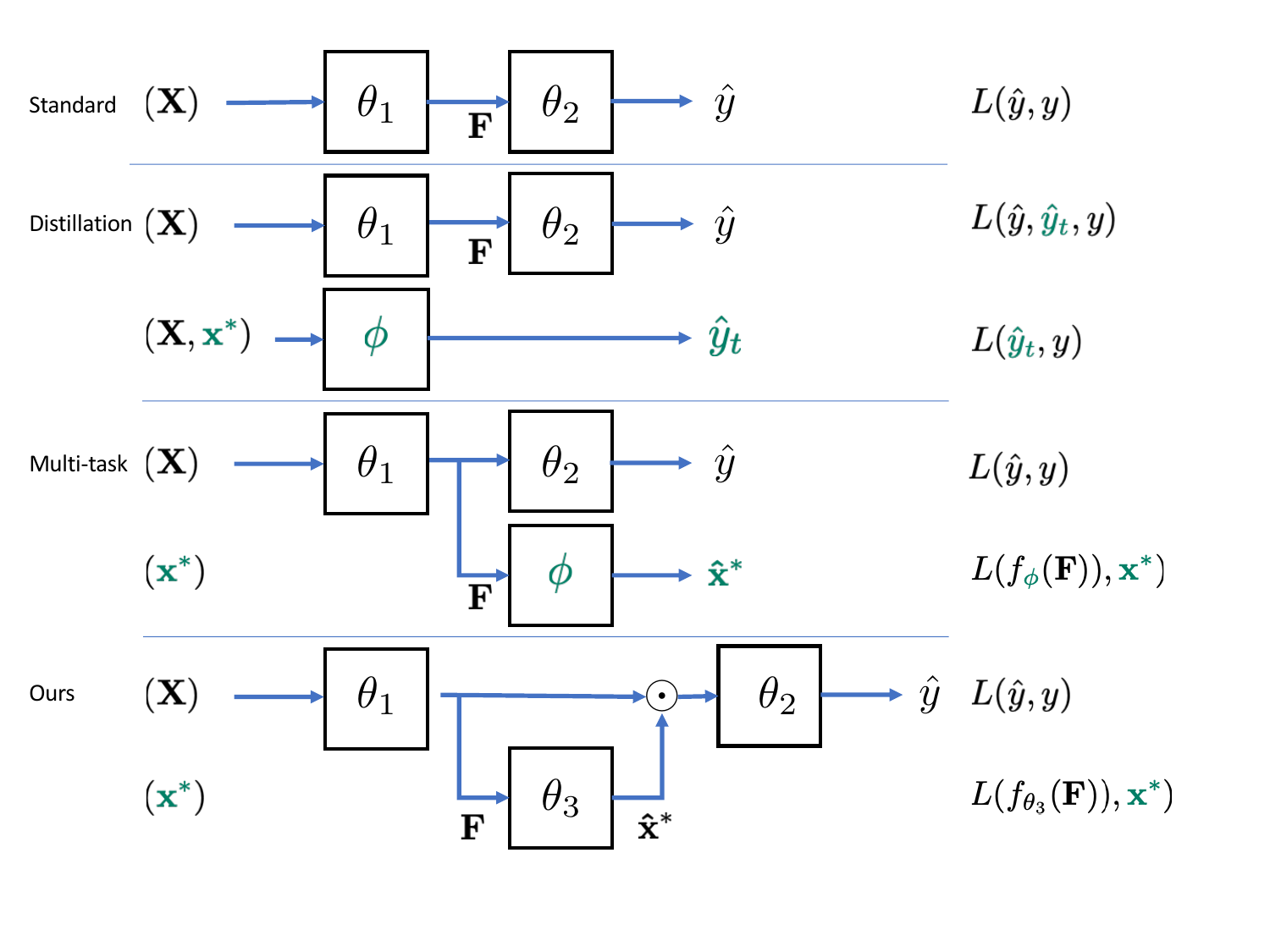}
    \caption{Comparison of learning strategies. In the standard network with parameters $\theta$, an input $\v{x}$ is mapped to a latent encoding $\v{F}$ and {then} on to a prediction $\hat{y}$.
    Distillation first learns a teacher network with parameters $\phi$ using also privileged information $\xstar$, then learns the weights $\theta$ to approximate that teacher network.
    Multi-task learning jointly learns to predict also $\xstar$ with a decoder with parameters $\phi$.
    The proposed framework adds an attention mechanism with parameters $\theta_3$ and supervises it with $\xstar$.
    \textcolor{PineGreen}{Green} denotes quantities used only during training.
}\label{fig:arch}
\end{figure}

{\textbf{Learning under Privileged Information}} attempts to leverage additional information $\xstar$ during training, but does not rely on it at test time, see Figure~\ref{fig:arch}. How to best exploit such side information is not obvious. Several algorithms have been developed for SVMs, for tasks including action~\cite{IB_visual2016} and image ~\cite{rank_Sharmanska_2013_ICCV} recognition. Applications in the context of deep learning include object detection~\cite{modalityhallucination_2016} and face verification~\cite{borghi2018face}.
Also simulated data has been interpreted as privileged information~\cite{lee2018spigan}, and (heteroscedastic) dropout has been used as a way of injecting, at training time, privileged information into the network~\cite{lambert2018deep}.

{\textbf{Knowledge Distillation (KD)}}\cite{hinton2015distilling,bucilua2006modelcompresion}, originally introduced for model compression, is closely related to the concept of privileged information~\cite{LopSchBotVap16}, see Figure~\ref{fig:arch}. KD trains a student network to imitate the output of (usually much bigger) teacher network pre-trained on the same task. To distil knowledge of both high- and low-level features from a pre-trained teacher, different variants of KD match feature maps at varying stages of the networks, usually to obtain more compact models~\cite{romero2014fitnets,jang2019learning,Zagoruyko2017AT}.

{\textbf{Multitask Learning}} could be viewed as a na\"ive way of incorporating privileged side information, by training an auxiliary task to predict the side information, see Figure~\ref{fig:arch}. The hope is that a shared feature representation will benefit the target task, because it profits from the additional supervision afforded by the auxiliary task. 
Examples in medical imaging \cite{pesce2019learning} and video description \cite{zhou2019grounded} use bounding boxes for such purposes.
%
There is a risk that tasks will instead compete for model capacity, leading to decreased performance. Several works focus on the non-trivial task of correctly balancing them~\cite{Sener_multitaskopt_2018,Kendall_2018_CVPR,similarity_grad}.

{\textbf{Few-shot Learning}} deals with the extension of an already trained classifier to a novel class for which there are only few examples.
The hope is that the new class, when embedded in the previously learned feature space, has a simple distribution that can be learned from few samples \cite{lake2015human}. One way to achieve this is to enforce compositionality of the feature space \cite{andreas2018measuring,tokmakov2018learning} by exploiting additional attributes of the training data, which can be seen as a form of privileged information.

\textbf{Pooling.} Virtually all image classification methods use some sort of pooling over a feature map extracted from a feature extractor -- nowadays a deep backbone. Beyond simple average- or max-pooling, other methods like bilinear pooling~\cite{kim2016hadamard,gao2016compact}, covariance pooling~\cite{Li_2018_CVPR,Li_2017_ICCV} and higher-order estimators \cite{cui2017kernel,cai2017higher} have been proposed. These methods are collectively referred to as \emph{second-order methods}, since they estimate second-order statistics of the features distribution. Empirically this can improve discriminative power~\cite{Li_2017_ICCV,cui2017kernel}.
Using a form of attention,
\cite{attentional_pooling_girdhar_NIPS2017} tackle action recognition using a bilinear pooling based on~\cite{kim2016hadamard,gao2016compact} and use one of the low-rank vectors to encode pose {as} privileged information into the network. Unlike ours, that approach is not applicable to first order pooling or to other forms of second-order pooling such as covariance pooling.

Despite recent developments in transfer learning and pooling, it is an open question how to leverage sparse, but highly informative privileged information at training time. We address this with a simple yet effective \emph{privileged pooling} scheme.

\section{Method}

\subsection{Background and Problem Statement}
Consider a supervised image classification task, with inputs $\v{X} \in \mathcal{X}$  represented as 3D tensors of size  $ W_{\v{X}}\times H_{\v{X}} \times 3$,
and outputs $y$ from a label space $\mathcal{Y}$.
The goal is to learn a function
$f_{\vg{\theta}}: \mathcal{X} \rightarrow \mathcal{Y}$ with parameters $\theta$, for instance a convolutional network (CNN), that minimises the expected loss $ l: \mathcal{Y} \times \mathcal{Y} \rightarrow \mathbb{R}$:
\begin{equation}
  \underset{{\vg{\theta}}}{\text{argmin}} \ \ \ 
 \mathbb{E}_{(\v{X},y) \sim P(\v{X},Y)} 
 \big[ l(f_{\vg{\theta}}(\v{X}),y) \big].
    \label{eq:empirical_risk}
\end{equation}

In the paradigm of learning under privileged information, we have access to additional side information denoted by $\v{x}^\star \in \mathcal{X}^\star$ for the training examples (but \emph{not} for the test data). {For the case of image classification, the supplementary information $\v{x}^\star$ typically has much lower dimension than the input image. Here, we consider annotated keypoint locations. In other words, } the training set is composed of triplets of the form $\{\v{X},\v{x}^\star,y\}$. 
As we will only have access to $\v{X}$ at prediction time, the overall goal is to minimise the risk of Eq.~\eqref{eq:empirical_risk}. However, we would want to leverage the information offered by $\v{x}^\star$ to regularise the training procedure. This leads to the new optimisation problem:
\begin{equation}
  \underset{{\vg{\theta}}}{\text{argmin}} \ \ \  
 \mathbb{E}_{(\v{X},y) \sim P(\v{X},Y)} 
  \big[ l(f_{\vg{\theta}}(\v{X}),y) \big] + g(\vg{\theta},p\big(\v{X},\v{x}^\star,y)\big),
    \label{eq:empirical_risk_reg}
\end{equation}
where $g$ represents a regulariser that depends on the learned parameters $\theta$ and on the joint distribution of the triplets $p(\v{X},\v{x}^\star,y)$. The challenge is to come up with an appropriate regulariser $g$ that alters the parameters $\theta$ in such a way that the generalization error for unseen data $\v{X}$ is reduced at test time. 
For many CNNs, $f_{\vg\theta}$ can be decomposed
into a feature extractor $f_{{\vg\theta}_1}(\v{X})$ that yields a feature map $\v{F}$ of size $ W\times H\times D$, followed by a (first-order) pooling operation $\pool(\v{F})$ that yields a feature vector $\v{p}$ of size $D$, and finally a multi-layer perceptron $f_{{\vg\theta}_2}$ that outputs a vector $\v{y}$ of class scores. {The spatial dimension of the feature map $\v{F}$ depends on the feature extractor, often it is smaller than the input image $\v{X}$. It is straight-forward to interpolate the input and/or the feature map so that their spatial dimensions match.}

\begin{figure}[t]
    \centering
    \includegraphics[width=\columnwidth, trim={0cm 0cm 6cm 0cm},clip]{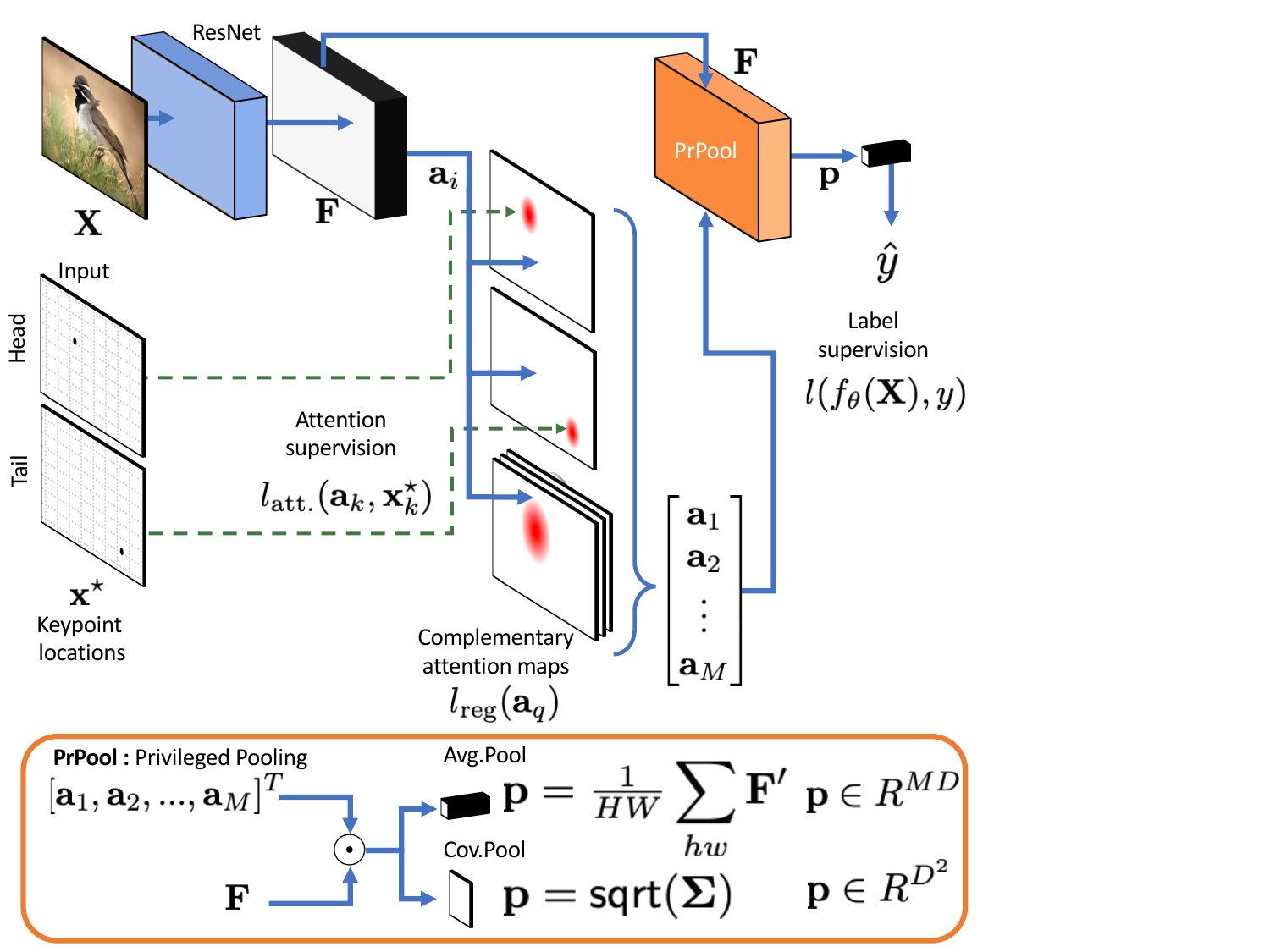}
    \caption{Privileged Pooling (\textbf{PrPool}) illustration.
    $M$ attention maps with $K$ supervised and $Q$ complementary ones. $F'$ is the extended feature map obtained using all attention maps (See Eq. \ref{eq:extended_feature_map}.)
    $\mathsf{sqrt}(\v{\Sigma})$ is the square-root normalized covariance matrix of the expanded feature map $F'$.
    \textcolor{PineGreen}{Green}-dotted lines denote quantities used only during training.
   }
    \label{fig:attention_mech}
\end{figure}

As discussed in Section~\ref{sec:related_work}, privileged information encompasses several forms of transfer learning, e.g., in the case of multi-task learning the second term of the objective function corresponds to $\mathbb{E}_{p(\v{x}^\star,\v{X})} \left[ l(g_{\vg{\theta}_1}(\v{X}),x^\star) \right]$, where $\vg{\theta}_1$ are the shared parameters of the common feature extractor.
This formulation, however, does not guarantee that the main task makes full use of the privileged information predicted for a test sample. Our architecture does exactly that: the attention maps can be seen as a strategy to pass the keypoint predictions learned from the privileged information back to the classification head, so as to highlight the most important visual features for animal species recognition. In this way we obtain a \emph{visual attention} mechanism~\cite{xu2015show,Lu2017CVPR} that steers the focus of the main network $f_{\vg{\theta}}$ towards locations in the image that contain important features for the classification task, see Fig.~\ref{fig:attention_mech}.

\subsection{Supervision of Attention Maps}

The purpose of attention mechanisms is to emphasize image evidence that supports prediction~\cite{selvaraju2017grad,tellmewheretolook_2018}.
In images this is commonly done by means of a $1\times 1$ convolution that outputs a weight for re-weighting features before passing them to the next network layer.
For example,~\cite{hu2019see,zheng2017learning} use attention maps to learn feature gating for fine-grained classification without additional supervision other than the image-level class label.
Here we explore a supervised attention mechanism: privileged information in the form of keypoint annotations is available at training time and serves to teach the network how to identify locations of interest in the latent feature representation.

As annotations we provide, for every training image, the desired output label as well as a set of $K$ keypoint locations. Keypoints are ordered and every point has a fixed semantic meaning, in our case a specific body part of the animal.
We found that scheme particularly effective for our application, as it delivers highly informative privileged information with fairly low annotation effort.

The framework is depicted in Figure~\ref{fig:attention_mech}. We add a network branch that derives $K$ attention maps $\v{a}_k$ from the feature map $\v{F}$. In contrast to previous approaches we rely on $3 \times 3$ convolutions to produce the attention maps.
This is necessary since we need to have a larger receptive field to produce attention maps that re-weight based on {surrounding pixels using} higher level concepts from the image (i.e. head, tail, etc.) instead of just the feature vector itself.

Keypoint locations can be represented in different ways. A simple idea would be a list of $K$ image coordinates $(x,y)$ that locate the keypoints in the image. A second possibility that better suits our approach is to create a set of $K$ binary maps with the same spatial dimensions as the input image, with pixels set to 1 at keypoint locations and 0 otherwise. This allows us to train our attention maps $\v{a}_k$ with a binary cross-entropy loss:
\begin{align}
    l_\text{BCE}(\v{a}_k,\v{x}^\star_k) = \frac{-1}{WH}\sum_{w,h} \big( x^\star_{whk} &\log(a_{whk}) \nonumber \\
     + (1 - x^\star_{whk}) &\log(1 - a_{whk})\big) \;,
    \label{eq:bce_loss}
\end{align}
{where $\v{x}^\star_k$ represents the binary map for the $k$-th keypoint location,
and $\v{a}_k$ is the predicted attention map with values in the continuous interval $[0,1]$. Note that the keypoint locations $\v{a}_k$ can easily be interpolated to fit feature maps $\v{F}$ of different resolutions, depending on the network architecture.}

Because the keypoint annotation might sometimes not be exactly at the right position, it is convenient to adopt a multi-scale loss. Attention map and keypoint map are passed through max-pooling operators $\{\mathcal{S}_1..\mathcal{S}_J\}$, with different kernel-sizes to account for different scales. Finally all the losses (Eq.~\ref{eq:bce_loss}) are combined:
\begin{equation}
    l_{\text{attention}}(\v{a}_k,\v{x}^\star_k) = \sum_{j}
    l_\text{BCE}(\mathcal{S}_j(\v{a}_k),\mathcal{S}_j(\v{x}^\star_k)).
    \label{eq:MS_bce_loss}
\end{equation}
The multi-scale attention loss $l_{\text{attention}}$ (Eq.~\ref{eq:MS_bce_loss}) is then applied separately to all $K$ keypoint maps.

{Note that the keypoints are not always all visible in the input image. If the $k$-th keypoint is missing, then $\v{x}^\star_k$ is set to 0 everywhere. This choice reflects the preference that, if a keypoint is not visible, the network should learn to predict its absence, corresponding to an empty attention map.}

\textbf{Complementary attention maps}
{are also included. Although they are not supervised by any keypoint annotation, they allow }
 the network to attend to potentially important regions not indicated by keypoints. For these additional attention maps,  {proper regularisation} is necessary, otherwise the optimization may converge to the trivial solution of a uniform map\remove{, or to multiple attention maps that attend to the same area}. The center loss \cite{wen2016discriminative} has been successfully used to enforce a single feature center per label and penalize distances from deep features to their corresponding center, see~\cite{hu2019see}. We empirically found that, in our case, a much simpler regularization that maximizes the variance within each attention map yields better results:
\begin{align}
     {\bar{a}}_q  = \dfrac{ \sum_{w,h}\v{a}_{whq}} {WH}, \ \ \ \ \ \ \ \ \ \ \ l_\text{reg}(\v{a}_q) = {\bar{a}}_q \cdot (1- {\bar{a}}_q).
    \label{eq:reg_loss}
\end{align}
In Eq.~\eqref{eq:reg_loss}, ${\bar{a}}_q$ represents the average value of the complementary attention map $q$. If that map is trivial (constant), then ${\bar{a}}_q$ will tend to either 0 or 1. A simple way to penalize extreme values is to maximize $l_\text{reg}(\v{a}_q)$. Intuitively, if we consider ${\bar{a}}_q$ as the parameter of a Bernoulli distribution, $l_\text{reg}(\v{a}_q)$ corresponds to its variance: a larger value translates to a more heterogeneous attention map.
The proposed regulariser ultimately imposes a bias against trivial maps that are overly diffuse (or, in the extreme case, uniform). {We have also tested various other, more complex, regularisations; but did not observe empirical improvements over the proposed one, see Appendix~\ref{sec:compl_loss} for details.}
%
The final loss for a model with a total of $M$ attention maps, including $K$ supervised and $Q$ {complementary} attention maps is defined as:
\begin{equation}
L = 
    l(f_\theta(\mathbf{X}),y) + \\
\frac{1}{K}\sum_k^K l_{\text{attention}}(\mathbf{a}_k,\xstar_k) - \\
\frac{1}{Q}\sum_q^Q l_\text{reg}(\mathbf{a}_q).
\end{equation}
    
\subsection{Attention Pooling}

Pooling operations integrate information over the spatial dimensions of a feature map $\v{F}$ with size $H\times W \times D$, to obtain a vector $\v{p}$, assumed to have the necessary representative power for image-level classification. 
{For average pooling, the $d$-dimensional vector $\v{p}$ is simply the per-channel mean of $\v{F}$:
\begin{equation}
    \v{p}^\text{AvgPool} = \frac{1}{HW} \sum_{h,w} \v{F}_{hw}.
\end{equation}
}
{Attention map serve to increase the representative power of $\v{p}$.}
To that end, we first expand the dimension of the feature map $\v{F}$ using the attention maps. We denote the expanded feature vector as $\v{F'}$ with size $H \times W \times M \times D$, computed as
\begin{equation}
    \v{F'}_{md} = \v{F}_{d} \odot \v{a}_{m}\;;
    \label{eq:extended_feature_map}
\end{equation}
with $\odot$ the Hadamard product over the spatial dimensions $h$ and $w$. The $M$ attention maps determine from which image regions features are be emphasized, respectively ignored.

This simple formulation allows the new feature map $\v{F'}$ to be used together with a range of pooling operations, including traditional first order pooling, low-rank bilinear pooling and covariance pooling approximations.

\textbf{First Order Pooling} comprises average and max pooling, the most common pooling operations.~\cite{Zhang_2019_ICCV} demonstrate that combining average and max pooling operations yields better results on the CUB200 dataset for fine-grained classification. 

Given the expanded feature map $\v{F'}$, we can take samples coming from the same attention map to perform $M$ average pooling operations: 
\begin{equation}
    \v{p}^\text{AvgPrPool} = \frac{1}{HW} \sum_{hw} \v{F'}.
\end{equation}
As the feature vectors $\v{p}_m^\text{AvgPrPool}$ are collected from different regions according to $\v{a}_m$, they preserve some degree of locality in the features . Note that total number of elements in the pooled representation is $DM$.

\textbf{Second Order Pooling}
regards each feature vector $\v{F'}_{hwm}$ in the expanded feature map $\v{F'}$ as a sample and computes a covariance matrix among the features. The mean and covariance matrix are computed by averaging over the spatial dimensions, $h$ and $w$, as well as over the attention map dimension $m$.
If the feature dimension $D$ is too large,  a $1\times1$ convolution on $\v{F'}$ can be used to reduce its size to $\Tilde{D}\ll D$, to save computational resources {as previously proposed by~\cite{Li_2018_CVPR}}. 
After reshaping the feature map $\v{F'}$ to $\Tilde{D} \times S$, where $S=HWM$, the covariance matrix can be computed as:
\begin{equation}
    \v{\Sigma} = \frac{1}{S}(\v{F'} - \v{\Bar{F'}})(\v{F'} - \v{\Bar{F'}})^T\;,
\end{equation}{}
with $\v{\Bar{F'}}$ the average feature value over the $S$ samples.
Furthermore,~\cite{Li_2017_ICCV} showed that normalising $\v{\Sigma}$ by taking its square root, denoted here as $\mathsf{sqrt}(\v{\Sigma})$, drastically improves the representation power of the features. 
%
An effective method consists in using the Newton-Schulz iterative matrix square root computation, which can be efficiently implemented on GPU~\cite{Li_2018_CVPR}.

Finally, note that one can backpropagate through the whole operation by using Newton-Schulz iterative square root computation, which enables end-to-end training of the network, with fine-tuning of the attention maps. We use the square root as our feature vector $\v{p}$ of size $\Tilde{D}^2$:
\begin{equation}
    \v{p}^\text{CovPrPool} = \mathsf{sqrt}( \mathbf{\Sigma} ).
\end{equation}

Other types of Second Order Pooling focus on a low rank approximation to perform bilinear pooling, although it can easily be computed in combination with the feature map $\v{F}'$ we observed that the best representative power came from the covariance pooling described above.
{Finally, note that covariance pooling is also possible over the original (not expanded) feature map $\v{F}$, leading to standard covariance pooling without attention.}

\section{Experiments}

We evaluate the proposed method on {three} different datasets, {showing} how it improves over Average Pooling and Covariance Pooling, and compare its performance against other methods that also leverage privileged information.

\subsection{Datasets}

\textbf{CUB200 \cite{WelinderCUB2010}} is a dataset of 200 different bird species, with a total of 5,994 training images and 5,794 test images. Each image comes with 15 keypoint annotations for body parts such as beak, belly, wings, etc.
CUB200 has been extensively used for fine-grained image classification and highly specialised architectures have been designed for it. Still, a vanilla ResNet-50 pretrained on ImageNet achieves 86\% accuracy (note that ImageNet includes some bird classes, and there might even be some common images between the two datasets).

Our focus is to evaluate keypoint annotations as privileged information to \emph{(i)} train a model in data-scarce settings, and \emph{(ii)} improve the generalization to other data.
Images in CUB200 are usually centered on the bird and depicted in ``standard" poses suitable for recognition, as in a field guide. To test generalization, we use images from the iNaturalist-2017 \cite{Horn_2018_CVPR} dataset, which features less curated, more challenging images of birds. 156 birds species are shared between CUB200 and iNaturalist-2017, for those species there are in total 3,407 images (average 22 samples/species) in iNaturalist-2017, which we use as an additional test set, termed \textbf{iBirds{2017}}.

{
\textbf{iBirds2018} is a subset of iNaturalist2018~\cite{Horn_2018_CVPR}, comprising 1258 bird classes with a total of 143,950 training and 3,744 validation samples. As this \textbf{iBirds2018} dataset exhibits a particularly long-tailed (realistic) class distribution, we separate it into different subsets according to the number $N$ of samples available per class. {First, as commonly done in the literature we consider 4 large sub-sets:} 308 many-shot ($N>100$), 563 mid-shot ($20\leq N\leq100$), 387 low-shot20 ($N<20$) and 188 low-shot15 ($N<15$). {Second, in a similar way, we also consider 9 sub-sets for a more fine-grained analysis.}
Furthermore we introduce \textbf{iBirds2018+}, where we annotated 5 samples in each low-shot15 class with keypoints. For consistency we used the same keypoints as in CUB200. One day of student work was required to write and use a small Python script for keypoint annotation to annotate 1,014 samples.}

\textbf{Caltech CameraTrap-20+ (CCT20+)} is a reduced version of CCT20\cite{Beery_2018_ECCV} augmented with privileged information.  CCT20 is a set of 57,000 images captured at 20 different camera locations and showing 15 different animal classes.%
\footnote{CCT20 is a subset of the bigger Caltech CameraTrap set with 243,000 images from 140 locations.} %
Images from 10 camera locations and taken on even days form the training set (13,000 samples).
There are two different test scenarios. The ``Cis" split (15,000 samples) consists of the odd days of the same cameras used for training, to test generalization across time for a fixed set of viewpoints. The ``Trans" split (23,000 samples) are images from camera locations not seen during training, to test generalization to new viewpoints.
As validation data, a single day (3,400 samples) for the Cis scenario, respectively a single location (3,400 samples) for the Trans scenario, is held out from the training data.

\emph{CCT20+} augments CCT20 with privileged information that we manually annotated for 1,182 images across all species and all cameras of the training set. We chose keypoints that have the same semantic meaning across the different species of animals in the dataset: head, left-front-leg, right-front-leg, left-back-leg, right-back-leg, tail and body-center.

In our experiments, we do not use the sequence information but treat every image independently. Moreover, we disregard images with more than one animal species and images without any annotated bounding box (the bounding box is not used in our system, we only use it as an indication that an animal was visible for the human annotator). Further details on the dataset annotation and statistics are available in Table~\ref{tab:cct_summary}.

\begin{table}[t]
 \begin{center}
    \begin{threeparttable}
    	\caption{Samples per Class in the CCT20 and CCT20+ (marked as Train+) Dataset After Disregarding Cars Class and Sequence Information}
       \begin{tabular}{l*{6}{R{0.8cm}}} 
 & \multicolumn{6}{c}{Datasplit}  \\ \cline{2-7}
 & \multirow{2}{*}{Train} & \multirow{2}{*}{Train+}  & \multicolumn{2}{c}{Val} & \multicolumn{2}{c}{Test} \\
Class &   &     &   Cis & Trans &    Cis &  Trans \\
\midrule
All      & 13,139 & 1,182 & 3,408 & 1,605 & 15,469 & 22,626 \\
opossum  &  2,470 &   178 &   346 &   425 &  3,988 &  4,614 \\
rabbit   &  2,190 &   189 &   320 &     9 &  1,461 &    669 \\
empty    &  2,122 &     0 & 1,860 &   192 &  3,922 &  6,355 \\
coyote   &  1,200 &    88 &   161 &    43 &  1,096 &  1,706 \\
cat      &  1,164 &   111 &   169 &    70 &  1,455 &  1,233 \\
squirrel &  1,024 &   109 &   146 &     0 &    496 &    779 \\
raccoon  &    845 &   105 &   116 &   108 &    869 &  4,314 \\
bobcat   &    673 &   110 &    92 &   624 &    751 &  1,901 \\
dog      &    580 &    77 &    89 &    80 &    627 &    631 \\
bird     &    353 &    78 &    36 &     5 &    360 &    127 \\
rodent   &    260 &    44 &    46 &     0 &    140 &     22 \\
skunk    &    212 &    73 &    24 &    49 &    162 &    263 \\
deer     &     38 &    12 &     2 &     0 &    136 &      0 \\
fox      &      5 &     5 &     0 &     0 &      2 &      1 \\
badger   &      3 &     3 &     1 &     0 &      4 &     11 \\
\end{tabular}

    	\label{tab:cct_summary}
  \end{threeparttable}
    \end{center}
\end{table}

\subsection{Implementation Details}

All experiments have been implemented in PyTorch \cite{pytorch_paszke2017}, with ResNet-101 pre-trained on ImageNet \cite{imagenet_cvpr09} as a backbone. \emph{WSDAN}~\cite{hu2019see} was implemented in Tensorflow \cite{tensorflow2015-whitepaper} and uses InceptionV3 as a backbone.
WSDAN experiments with ResNet-101 did not yield comparable results, which is why we included WSDAN with its original backbone. See 
the Supplementary material for more details. The results from WSDAN differ slightly ($<1.6$\%) from the 89.4\% accuracy on CUB200 reported in the original publication, {note that, in contrast to the original publication, we present an average of different random seeds using original code.}
All our models are trained with SGD with momentum 0.9 and weight-decay 10$^\text{-4}$, batch-size 10 and initial learning-rate of 0.01, a backbone multiplier of 0.01, and exponential decay by a factor 0.9 every 1,000 iterations. 

Some methods, such as~\cite{hu2019see}, leverage the predicted attention maps to compute a bounding box around the attended area, crop the image and re-feed it through the network. The final output is obtained by averaging the two  predictions. This general strategy is orthogonal to what we propose in this paper and can be applied also with methods that predict attention maps.
This practice is sometimes beneficial, but not always, depending on the dataset. We report results for 
{all datasets with re-feeding, except for CCT.}
See the ablation study in Section \ref{sec:crop_testtime} for an analysis of this practice.

 \begin{table}[t]
 \begin{center}
    \begin{threeparttable}
 	\caption{Comparison of Model Size}
         \begin{tabular}{lccc}
Model &  FLOPS [G] &  Params [M] &  Time [ms]  \\
\hline
Avg.Pool   &       39.9 &        45.0 &       15.6        \\
Avg.PrPool &       48.3 &        68.3 &       19.5    \\
WSDAN      &       40.0 &        57.7 &       17.9 \\
S3N*       &       90.1 &       104.3 &       47.1    \\
Cov.Pool   &       40.3 &        49.6 &       16.6         \\
Cov.PrPool &       84.3 &        52.0 &       36.8    \\
TransFG    &       107.56 & 86.85 & 116.28 \\
\end{tabular}

\begin{tablenotes}
\item {CUB200 default hyperparameters, ResNet-101 as backbone for all baselines except TransFG. $^*$: s3N can only use ResNet-50, see text. The displayed times are average inference times on a Nvidia TitanX GPU}
\end{tablenotes}
 	\label{tab:model_size}
\end{threeparttable}
\end{center}
\end{table}

\subsection{Baselines}

\subsubsection{No-$\xstar$ Methods} 
As baselines we consider several recent works that pretrain only on ImageNet, and whose network architecture has a complexity similar to ours.

\emph{AvgPool}: is a vanilla Resnet-101 pre-trained on ImageNet, with average pooling and a single fully connected layer for final prediction.

\emph{WSDAN Avg}~\cite{hu2019see} is a method that reaches state-of-the-art results on several fine-grained image classification datasets including CUB200~\cite{WelinderCUB2010} FGVC-Aircraft~\cite{maji2013fine}, Stanford Cars \cite{Krause_2013_ICCV_Workshops} and Stanford Dogs \cite{khosla2011novel}. It uses unsupervised learning of attention maps to perform weighted pooling.

iSQRT~\cite{Li_2018_CVPR} (\emph{CovPool}) proposes to change the usual average pooling of feature maps with a square-root normalized covariance pooling. Empirically this is particularly useful for fine-grained classification, where 2$^\text{nd}$-order information can be highly informative. {The dimension of the feature map is always reduced from 2048 to 256 to lower the computational cost, in all our experiments.}

S3N~\cite{ding2019selective} proposes a way to select peaks of the feature response, so as to force the network to explore those peaks, which can be especially informative for the prediction. That method also achieves state-of-the-art performance on CUB200, but it does so at a high cost: $d^2$ additional parameters are required for the peak sampling layers from a feature map of size $W\times W\times D$. {S3N is the only method that does not use ResNet-101 in its original implementation. To stay within GPU memory limits, it must be used with ResNet-50, because of its large number of additional parameters.}

{
LSTM~\cite{ge2019weakly} uses a weakly supervised object detector to detect different relevant parts of the image, and in a second step all these detected objects are feeded as a time-sequence of images to a LSTM to finally perform a prediction of an object class.}

{
TransFG~\cite{he2021transfg} is transformer-based network tuned for fine-grained classification, it uses an overlapping window to create patches over the image to select and discriminative image regions.}

\subsubsection{$\xstar$ Methods}

\begin{figure*}[t]
    \centering
    \includegraphics[width=\textwidth]{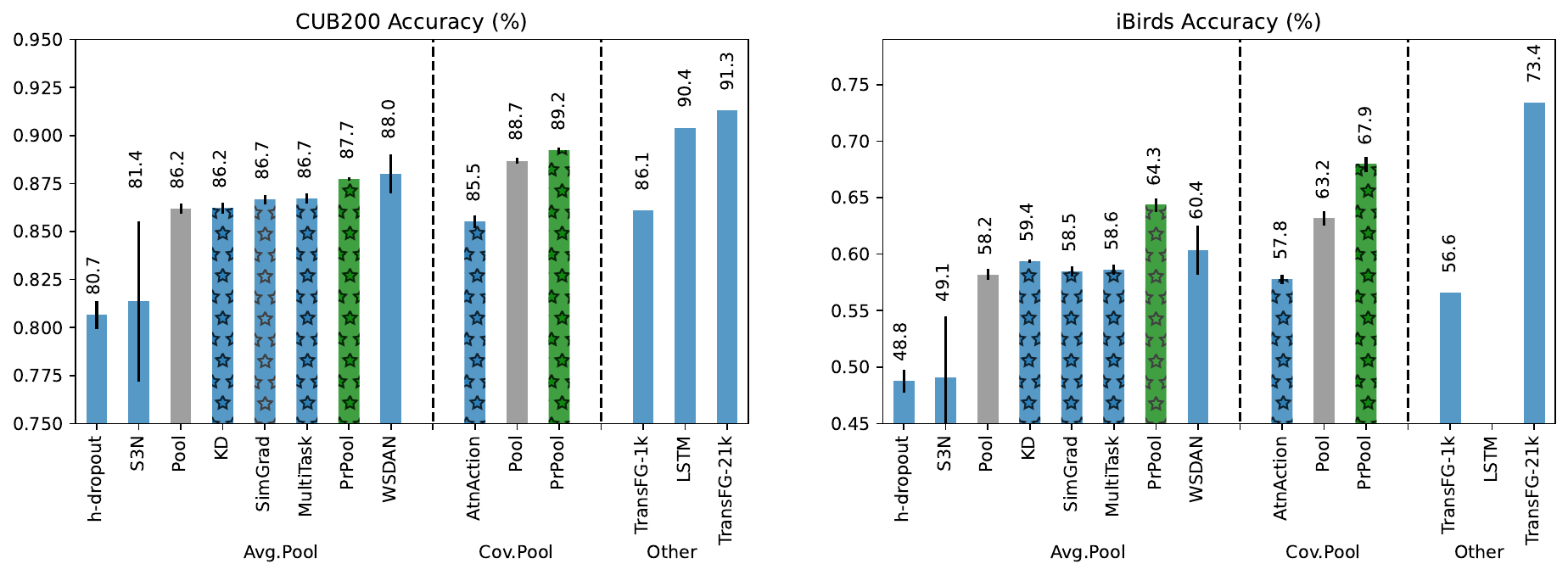}
    \caption{Top-1 accuracy for CUB200 (left) and iBirds{2017} (right) test datasets. {Mean accuracy and standard deviation (error bars) over 5 runs.} In gray the baselines methods \emph{AvgPool} and \emph{CovPool}; green indicates our methods trained with \emph{PrPool} (ours). Bars with $\star$ indicate use of privileged information at training time {($\xstar$ methods)}. LSTM values are taken directly from \cite{ge2019weakly}, TransFG-21k uses \cite{he2021transfg} pretrained on ImageNet-21k and TransFG-1k is pretrained on ImageNet-1k}
    \label{fig:cub_full}
\end{figure*}

We test our method \emph{PrPool} in combination with two different options: \emph{AvgPrPool} is a first order alternative where we compute mean and max pooling operations. \emph{CovPrPool} is a second order method using covariance pooling as described in the methods section {and a 1x1 convolution to reduce the dimensionality of the feature map from 2048 to 256}. For CUB200 {we follow WSDAN and use a total of 32 attention maps (in our approach, 15 are supervised by keypoints, the remaining 17 are complementary).}
For CCT20 we used {a total of 8 attention maps} (7 supervised and 1 {complementary}).

The \emph{Multitask} architecture is identical to \emph{AvgPool}, except that the output of the backbone is also connected to a fully connected layer that predicts the keypoints locations. This baseline represents a sort of ``lower bound" for the impact of privileged information that is available only during training.

\emph{SimGrad}~\cite{similarity_grad} is an improvement of the multitask architecture that aims at reducing the risk that the auxiliary task harms the main task. After separately computing the gradients of the two tasks w.r.t.\ the shared parameters, the gradients are averaged only if the cosine similarity between them is positive. Otherwise the auxiliary loss is ignored, with the  intuition that it should not influence the fitting if it is in conflict with the primary loss. 

For knowledge distillation \emph{(KD)} we train a classification network that has two input channels, one for the RGB images and one for the keypoint masks. Once trained, we distill the output of that teacher network into our baseline Resnet101 as student model.

Heteroscedastic dropout (\emph{h-dropout})~\cite{lambert2018deep} highlights how learning under privileged information can be implemented via a dropout regularization.
{We implemented this method in its original form and} in conjunction with other attention-based methods (e.g., \cite{hu2019see}), but found that, despite our best effort, the noise injected into the fully connected layer made training {unstable for masked inputs with bounding boxes (as in the original implementation), and also with masked inputs around keypoints with different diameters. We show results with the keypoint annotation version, because it empirically performed better}.

We have also tested Attentional pooling for action recognition (\emph{AtnAction})~\cite{attentional_pooling_girdhar_NIPS2017} on our datasets. This method uses a low-rank bilinear approximation and takes one of the low-rank vectors as attention maps to encode the privileged information.
{By default that method sets the rank to $L=768$, as we did not observed any significant effect when using larger values for $L$, so we kept the original value in all our experiments.}

A main goal of our work is to learn classes for which we only have few training examples, so it is also related to few-shot learning.
We test compositional \emph{FewShot} recognition~\cite{tokmakov2018learning}, which uses class-level labels to enforce compositionality (see Section~\ref{sec:related_work}). We use the same 5 random splits into base classes and novel classes as~\cite{tokmakov2018learning} and run our \emph{PrPool} network on them, uniformly sampling from the batches when creating a batch, in order to deal with the imbalance between novel and base classes.

In Table \ref{tab:model_size} we report the 
{ FLOPs, number of parameters and inference time for}
different baseline methods as well as for the proposed \emph{PrPool} model. As can be seen the overhead needed by \emph{PrPool} is limited compared to other methods such as \emph{S3N} or \emph{TransFG}. {We have used the original implementations of \textit{WSDAN}, \textit{CovPool}, \textit{S3N}, \textit{h-dropout} {and  \textit{TransFG}}; and our own re-implementations of all other methods. All presented results are from experiments we ran ourselves, unless stated otherwise.} { \emph{LSTM} could not be included in this analysis as there is no open code. It is also not directly comparable conceptually because it uses a recurrent network over a sequence of images in two different steps.}

\subsection{Fine-grained Classification}

Figure~\ref{fig:cub_full} presents the results on the CUB200 test set. The numbers are averages over 5 runs with different random initialisations. The average pooling baseline (\emph{AvgPool}) achieves 86.2\%. With average pooling supervised by privileged information (\emph{PrPool}), this increases to 87.7\%. Among the first order pooling methods in this case the best performance is achieved by \emph{WSDAN} with 88.0\%

In line with the literature~\cite{Li_2018_CVPR} we find that covariance pooling is superior to average pooling for fine-grained-classification, reaching {88.7\%}. The best \remove{overall} {result with pooling methods} for the CUB200 dataset is achieved by the proposed privileged pooling, which improves the result to {89.2\%}.
These improvements may seem comparatively small. When, however, testing the trained networks on {iBirds2017}, the gains are amplified, i.e., our models with supervised attention generalize better. The relative improvements are quite significant, up to {16.7\%} over the average pooling baseline.
%
While most methods show a mild improvement over the baseline without attention, they do not reach the performance of \emph{PrPool}. This includes other methods that also use privileged information at training time, but do not show a significant improvement on CUB200 and even fall behind the baseline in {iBirds2017}. Seemingly they are to some degree overfitted to the CUB200 distribution and not able to generalize.

{For completeness, we also include architectures with neither ResNet-101 nor pooling mechanisms that have recently demonstrated high performance on CUB200.}
%
{\emph{LSTM}~\cite{ge2019weakly} is based on a two stage prediction scheme and reports 90.4\% top-1 accuracy on CUB200. We could not evaluate that method on iBirds{2017} nor its time-complexity because no code is available (and accurate re-implementation is not possible given the description in the paper).}

More recently, TransFG~\cite{he2021transfg}, a transformer-based network tuned for fine-grained classification, was able to achieve 91.7\% accuracy on CUB200.
The network is pretrained on ImageNet-21k instead of the usual 1k classes which, by itself, is likely to improve transfer learning \cite{ridnik2021imagenetk}.
Beyond the sheer number of classes, ImageNet-21k contains \textgreater 350,000 samples from \textgreater350 different bird species, including \textgreater60 species also present in CUB200. This overlap suggests that pre-training on ImageNet-21k gives a substantial advantage when processing CUB200 that previously may have been underestimated. We found that TransFG indeed did not perform well when pretrained only on ImageNet-1k {as can be observed under category "Other" in Figure~\ref{fig:cub_full}.} A detailed analysis can be found in Appendix~\ref{sec:compl_transformer}.

\begin{table}[t]
	\begin{threeparttable}
	\caption{Top1 Accuracy for CUB200 and iBirds{2017} test set}
       \begin{center}
        \begin{tabular}{llC{0.5cm}C{0.5cm}C{0.5cm}rC{0.5cm}C{0.5cm}C{0.5cm}}
    & & \multicolumn{3}{c}{iBirds2017} && \multicolumn{3}{c}{CUB200} \\ \cline{3-5} \cline{7-9}
\multicolumn{2}{c}{Samples per class}      &         5  &   10 &   15 &&      5  &   10 &   15 \\
\midrule
\multirow{6}{*}{Avg} 
    & S3N 
    &       38.8 & 49.6 & 54.0 &&    67.2 & 79.6 & 82.4 \\
    & Pool &       39.3 & 49.1 & 54.6 &&    66.0 & 77.9 & 81.7 \\
    & SimGrad 
    &       34.6 & 47.4 & 53.2 &&    62.7 & 77.7 & 81.8 \\
    & MultiTask &       33.6 & 47.2 & 52.5 &&    61.7 & 77.5 & 81.5 \\
    & WSDAN 
    &       38.5 & 51.8 & 56.0 &&    68.8 & 80.4 & 84.1 \\
    & PrPool (ours) &      \underline{48.5} &  \underline{58.1} &  \underline{60.3} &&     \underline{73.8} &  \underline{82.5} & 84.6 \\
 \clineh{2-9}
\multirow{2}{*}{Cov} & AtnAction 
& 38.4 & 49.1 & 53.6 && 	68.7 &	78.1 &  81.6 \\
 & Pool 
 &       45.5 & 56.2 & 58.8 &&    72.0 & 82.4 &  \underline{85.0} \\
    & PrPool  (ours) &      \textbf{49.6}& \textbf{59.6} & \textbf{62.8} &&    \textbf{74.6} & \textbf{83.3} & \textbf{85.4} \\
\end{tabular}

    \end{center}
    \begin{tablenotes}
  \item  Best performance in \textbf{bold}. Second best \underline{underlined}.
    \end{tablenotes}
	\label{tab:cub_nshot}

   \end{threeparttable}
\end{table}

\begin{figure}[t!]
    \centering
    \includegraphics[width=\columnwidth]{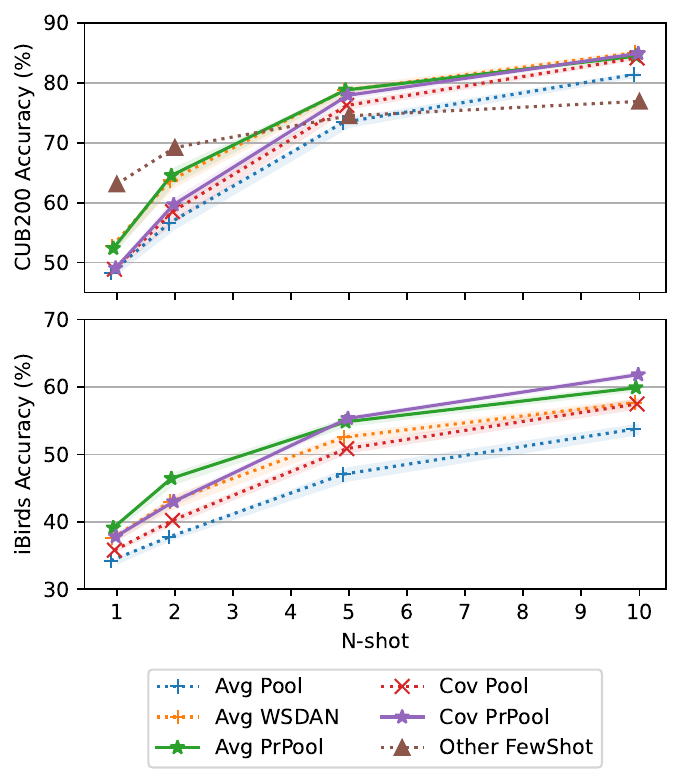}
	\caption{CUB200 (top) and iBirds{2017} (bottom) Top1 Accuracy. 100 base classes, with 100-way new classes with $n$-shots. Average over five random novel/base splits {with standard deviation as shaded areas}. Methods with {\large $\star$} marker denote PrPool (ours). FewShot results {taken directly} from~\cite{tokmakov2018learning}}
    \label{fig:n_shot_rep}
\end{figure}

\subsection{Data Efficiency}

Privileged information especially improves the data efficiency in data-scarce regimes. 
{We consider two main scenarios for experimental evaluation: a few-shot learning scenario with CUB200 (as done previously in the literature) and a long-tailed scenario with iBirds2018.}

{\textbf{Few-shot learning.}} We draw $n$ samples per class out of the {CUB200 training samples.}
Table~\ref{tab:cub_nshot} shows that \emph{PrPool} consistently outperforms all competing approaches, with increasing benefits as the training set gets smaller. Note that in the most challenging 5-shot case, there are only 1000 samples to learn 200 classes.
We also find that in the small data regime, the na\"ive multitask loss does not improve performance, and also other baselines become rather inconsistent.

\begin{figure*}[t]
    \centering
    \includegraphics[width=\textwidth]{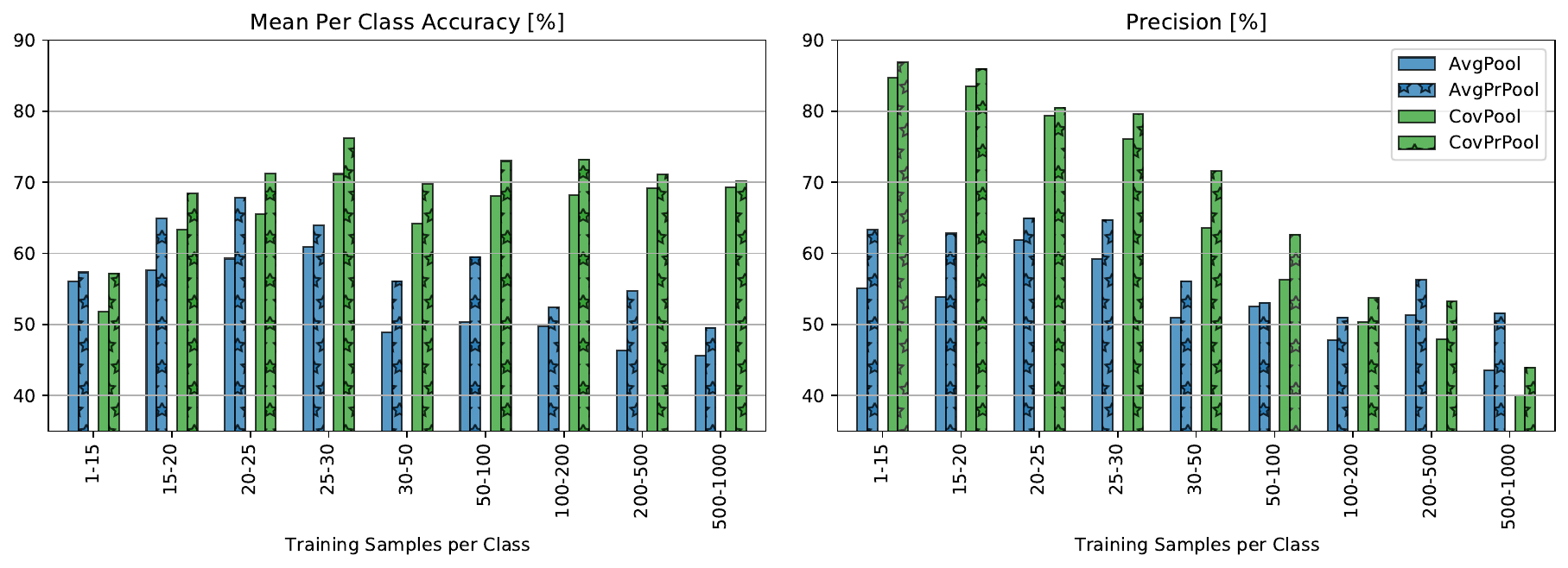}
	\caption{{Mean per class accuracy and precision for iBirds2018. Results are grouped in sub-sets according to the number of available training samples for each class.}}
    \label{fig:i_birds_detail}
\end{figure*}

Given the performance achieved with only 5 samples per class, we also compare to few-shot learning. We use the same evaluation strategy of~\cite{tokmakov2018learning} with 100 base classes and 100 new classes with only $n$ shots each. Results are displayed in Figure~\ref{fig:n_shot_rep}.
As expected, dedicated few-shot learning -- based on a set of well-trained base classes and some form of distance learning to add the new classes -- is superior in the extreme 1-shot and 2-shot scenario. But already in the 5-shot case, we find that even simple average pooling is competitive with few-shot learning, and our privileged pooling already outperforms it. At 10 samples the difference is accentuated, as one moves further away from the extreme few-shot setting.
Apparently the privileged information can, already at this low sample number, compensate the reduced sampling of the pose and appearance space, by steering the learning towards sub-regions with a well-defined semantic meaning across classes.

As before we evaluate the implemented methods on the iBirds{2017} dataset, see Figure~\ref{fig:n_shot_rep}. Results are consistent with the previous ones, \emph{PrPool} proves to be very effective at increasing performance in low-data regime and improves the generalization power of the network. {See Appendix~\ref{sec:compl_fewshot} for more baseline comparisons.}

{
\textbf{Long-tailed dataset.} In general, our method is not specifically designed to deal with long-tailed class distributions. We still evaluate on such a scenario nonetheless using \textbf{iBirds2018} to understand its impact on classes with scarce training labels.
Table~\ref{tab:ibirds2018} shows the top-1 and top-5 performance for the whole dataset and for different subsets with progressively lower number of training samples. Generally, \textit{CovPrPool} seems to best leverage the privileged information $\xstar$, reaching a top-1 accuracy of 68.8\%, 4 points higher than the best baseline method without $\xstar$. Importantly, the privileged information available boosts performance for the considered sub-groups. Figure~\ref{fig:i_birds_detail} shows Mean per Class Accuracy and Precision for more fine-grained sub-sets according to the number of training samples available. We observe that leveraging privileged information boosts precision consistently for all class sub-sets.
{It also becomes clear that \textit{CovPool} and \textit{CovPrPool} yield higher class accuracy than \textit{AvgPool} approaches on classes with $>$ 15 training samples; for classes with $>$ 200 training samples \textit{CovPool} and \textit{CovPrPool} also have higher accuracies but at the cost of a reduced precision (Figure~\ref{fig:i_birds_detail}, left).}
This effect is somewhat reduced for lower-shot classes though. \textit{CovPrPool} outperformed consistently all other forms of pooling considered in Figure~\ref{fig:i_birds_detail}. 
These experiments, on a highly popular dataset, demonstrate performance of our method for a realistic case where the label distribution is not uniform. Our method shows to be effective for cases with few labels at training time. The results support our hypothesis that a moderate labelling effort -- a handful of keypoints in a small subset of training images -- does lead to a significant performance boost.
%
}

\begin{table}[t]
 \begin{center}
    \begin{threeparttable}
 	\caption{{iBirds2018 Results}}
\resizebox{\linewidth}{!}{\begin{minipage}{\linewidth}
         \begin{tabular}{llrrrrrr}

 \multirow{2}{*}{Pool}    &    \multirow{2}{*}{Model} & \multirow{2}{*}{Top-1} & \multirow{2}{*}{Top-5} & \multicolumn{4}{c}{$N$-shot subsets}\\\cline{5-8}
 &  &   &   &  Many &  Mid &  Low &  Low15\\
\midrule
\multirow{2}{*}{Avg} & S3N &   47.9 &   71.5 &  50.8 & 49.0 &   43.9 &   41.1 \\
    & Pool &   54.0 &   77.9 &  47.4 & 55.7 &   56.7 &   55.1 \\
    & MultiTask &   58.9 &   80.7 &  69.5 & 58.8 &   50.5 &   47.2 \\
    & WSDAN &  \underline{64.8} &   84.5 &  \textbf{76.2} & 65.5 &   54.6 &   51.0 \\
        & PrPool (ours) &    59.5 &   81.0 &  52.6 & 62.6 &   \underline{60.3} &   \textbf{57.4 }\\\clineh{2-8}
\multirow{2}{*}{Cov} & Pool &  64.1 &   \underline{85.0} &  68.8 & \underline{67.4} &   55.6 &   50.5 \\
    & PrPool (ours) &   \textbf{68.8} &   \textbf{87.8} &  \underline{71.6} & \textbf{72.5} &   \textbf{61.2} &   \underline{55.7} \\
\end{tabular}

\end{minipage}}
\begin{tablenotes}
\item {Best performance in \textbf{bold}. Second best \underline{underlined}}
\end{tablenotes}
 	\label{tab:ibirds2018}
\end{threeparttable}
\end{center}
\end{table}

\subsection{Generalization with biased datasets}

\begin{figure}[h]
    \centering
    \includegraphics[width=\columnwidth]{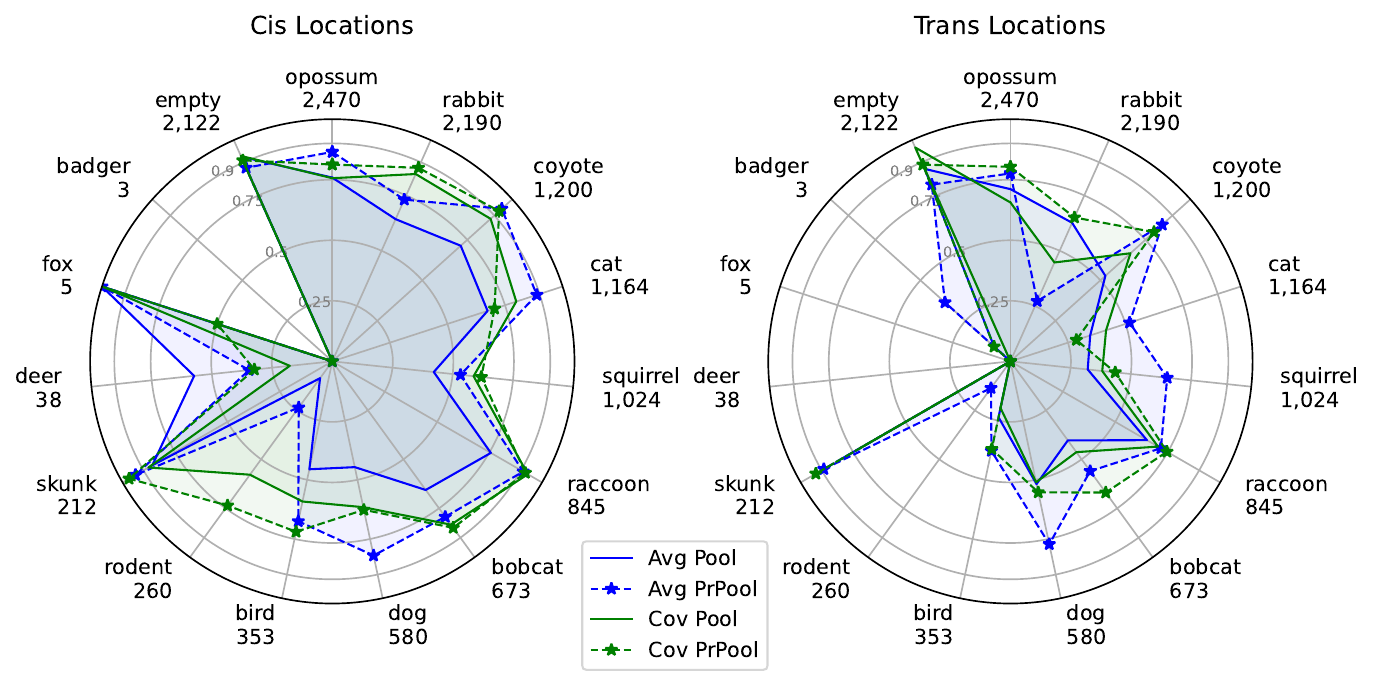}
	\caption{Accuracy per class in CCT20 test datasets. Training with CCT20: 13K samples including 1,180 samples with keypoint annotations. {Classes sorted according to number of training samples per class (indicated below each animal class). Methods with {\large $\star$} marker denote PrPool (ours).}
 }
    \label{fig:cct}
\end{figure}

The CCT20 dataset is very challenging, due to bad illumination, frequent occlusions, camouflage and extreme perspective that arise in camera traps. Moreover, the highly repetitive scenes are an ``invitation to overfit" and learn spurious correlations, which then hinder generalization to new scenarios (e.g., unseen camera locations). Our \emph{PrPool} method achieves the best performance. In this case, with fewer and more distinct classes, first-order pooling works better than \emph{CovPool} (but also the latter outperforms the baselines). 
{Figure \ref{fig:cct} shows that \emph{AvgPrPool} outperforms other types of pooling in several classes, such as squirrel, cat and dog.  
Table \ref{tab:cct_results} shows the results on CCT20. 
{In Cis locations the Mean per Class Accuracy has a slight increase with AvgPool methods when using the CCT dataset (13k training samples) instead of CCT+ (1k training samples): \emph{AvgPool} improves 2.8 points, \emph{AvgPrPool} improves 0.9 points. CovPool methods, on the other hand, have a larger improvement. \emph{CovPool} improves by 9.5 points while \emph{CovPrPool} improves by 12.9 points.
These findings are aligned with earlier results we observed on iBirds2018 where \emph{CovPool} methods yield overall better results with more training samples than \emph{AvgPool} methods}.
 We also find that attention-cropping at test time decreases performance for both \emph{WSDAN} and \emph{PrPool} for CCT datasets and therefore omitted that step. Please refer to Section \ref{sec:crop_testtime} for an ablation study on this effect.}

Note that our method trained only on the CCT20+ subset with $\approx$1000 samples outperforms the \emph{AvgPool} baseline even when the latter is trained with 10$\times$ more samples. We see two reasons for this, \emph{(i)} the superior data efficiency through Privileged Pooling, and \emph{(ii)} the low diversity of samples in Camera Trap data, where more samples can in fact reinforce inherent dataset biases.
In the most challenging Trans-location setting \emph{PrPool} reaches 75\% test set accuracy.

\begin{table}[tb]
  \begin{center}
	\begin{threeparttable}
	\caption{Overall Accuracy and Mean per Class Accuracy Results of Models Trained on CCT20 and CCT20+}
        \begin{tabular}{lllC{0.8cm}C{0.8cm}lC{0.8cm}C{0.8cm}}
&    & {} & \multicolumn{2}{c}{Acc} && \multicolumn{2}{c}{Acc\textsubscript{class}}  \\ \cline{4-5} \cline{7-8}
 &   & Dataset &    CCT+  & CCT && CCT+ & CCT    \\
     \multicolumn{8}{l}{\textit{Cis Locations}} \\ \clineh{1-8}
&\multirow{5}{*}{Avg} 
    & S3N &      75.2 &  75.5 &&     {61.7} &  66.7   \\
    &   & Pool &      71.4 &  73.6 &&      57.0 &  59.8         \\
   & & MultiTask &      72.2 &  76.4 &&      59.5 &  65.8         \\
 && WSDAN &      72.9 &  76.2 &&      52.2 &  63.9   \\
    &  & PrPool  (ours) & \textbf{81.0} & 	\textbf{83.5}&& 	\textbf{69.4}& 	\textbf{70.3} \\
       \clineh{2-8}
&\multirow{2}{*}{Cov} & Pool &      74.3 &  81.4 &&      59.8 &  69.3   \\
 &   & AtnAction &    72.7 & 	77.4 && 	56.3 & 	68.0   \\
 &   & PrPool (ours) &      {76.7} &	{82.9}&& 	57.0& 	{69.9} \\
 \multicolumn{8}{l}{\textit{Trans Locations}} \\ \clineh{1-8} 
&\multirow{5}{*}{Avg} 
 & S3N &         64.2 &  70.6 &&      40.2 &  50.9 \\
&& Pool &           65.1 &  66.1 &&      42.0 &  43.6 \\
&& MultiTask &           65.6 &  67.2 &&      42.6 &  50.0 \\
&& WSDAN &        62.6 &  61.4 &&      34.7 &  42.4 \\
&& PrPool (ours) &  \textbf{70.6} & 	72.3 && 	\textbf{52.7} & 	\textbf{54.8} \\
     \clineh{2-8} 
&\multirow{2}{*}{Cov} & Pool &            67.2 &  70.1 &&      41.9 &  43.8 \\
 &   & AtnAction &    64.4 & 	{73.0} && 	{43.2} & 	51.2  \\
 &   & PrPool (ours) &       {68.5} & 	\textbf{75.0}&& 	42.8& 	{51.6} \\

\end{tabular}

	 \label{tab:cct_results}

\begin{tablenotes}
 \item CCT dataset has 13K samples, including 1K samples with keypoints annotations. CCT+ only has 1K, all of them with keypoints annotations. Best performance in \textbf{bold}
    \end{tablenotes}
   \end{threeparttable}
    \end{center}
\end{table}

\begin{figure}[t]
    \centering
    \includegraphics[width=\columnwidth]{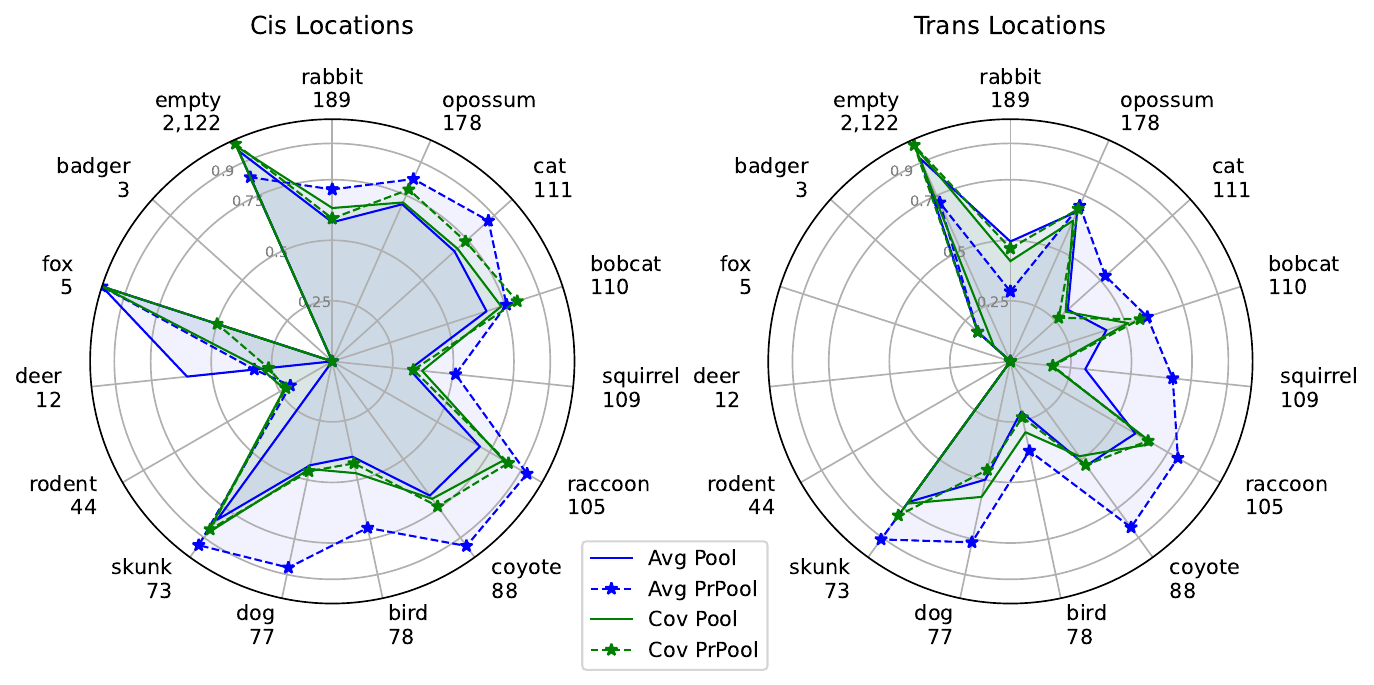}
	\caption{Accuracy per class in CCT20 test datasets. Training with CCT20+: 1,180 samples, all training samples are provided with keypoint annotations. 
 {Classes sorted according to number of training samples per class (indicated below each animal class). Methods with {\large $\star$} marker denote PrPool (ours).}
 }
    \label{fig:cct_plus_by_n}
\end{figure}

We now explore the impact that privileged information has when using only the keypoint annotated images from CCT20+ dataset. As expected, the differences in performance with respect to the baseline methods are larger in this case. See Figure~\ref{fig:cct_plus_by_n} for more details on the per class performance.

\subsection{Ablation study: attention maps}

We go on to analyze how important attention map supervision is in our architecture, in combination with both average and covariance pooling. To that end we train exactly the same architecture as \emph{PrPool}, but without the supervision signal $l_{\text{attention}}$ from keypoint annotations. The results in Table~\ref{tab:cub_supervision} confirm that the privileged information plays an important role and significantly increases prediction performance. Moreover, we observe that already the regulariser alone improves over totally unconstrained self-attention, as expected.

\begin{table}[tb]

  \begin{center}
	\begin{threeparttable}
	\caption{Top-1 Accuracy Results of Attention Maps With Different Supervisions.}
           \begin{tabular}{lll*{3}{C{0.5cm}}l}
&&& \multicolumn{3}{c}{Supervision Type} \\\cline{4-6}
~\rule{0.5in}{0in}& & &         No &  Reg &   Pr &\rule{0.5in}{0in}~\\   
&\multicolumn{5}{l}{\textit{iBirds2017}} \\ \clineh{2-6} 
&&AvgPool &       60.3 & 61.7 & 65.2  \\
&&CovPool &       62.3 & 66.5 & 67.0 \\
&\multicolumn{5}{l}{\textit{CUB200}} \\  \clineh{2-6}
&&AvgPool &          84.3 & 87.0 & 88.1 \\
&&CovPool &         87.1 & 88.5 & 89.0 \\
\end{tabular}

     \begin{tablenotes}
  \item \textbf{No} Supervision, \textbf{Reg}ularized Supervision ($l_{\text{reg}}(\v{a}_q)$)  and \textbf{Pr}ivileged Supervision ($l_{\text{attention}}(\mathbf{a}_k,\xstar_k)$)
  \end{tablenotes}
  \label{tab:cub_supervision}
  \end{threeparttable}
  \end{center}
\end{table}


{We also vary the number of attention maps used for the CUB200 dataset, see Table~\ref{tab:cub_nattention}. As it can be seen from the table, performance tend to slightly increase when complementary attention maps are added to the model, however too many might lead to a minor model overfit.
}

\begin{table}[t]
  \begin{center}
	\begin{threeparttable}
	\caption{{Top-1 Accuracy when Training With Different Numbers of Attention Maps.}}
           \begin{tabular}{lllrrrrr}
~\rule{0.4in}{0in}&& & \multicolumn{4}{c}{N. Attention Maps}&~\rule{0.4in}{0in} \\\cline{4-7} 
 &      &  &   15 &   16 &   32 &   64 \\
   &    \multicolumn{5}{l}{\textit{iBirds2017}} \\    \clineh{2-7} 
 && AvgPool & 64.0 & 64.2 & 64.3 & 63.8 \\
   &    & CovPool & 62.5 & 66.8 & 67.9 & 67.2 \\
   &\multicolumn{5}{l}{\textit{CUB200}} \\    \clineh{2-7} 
 && AvgPool & 87.2 & 87.1 & 87.7 & 87.5 \\
  &     & CovPool & 85.6 & 88.6 & 89.2 & 88.9 \\
\end{tabular}

     \begin{tablenotes}
   \item In all cases 15 attention maps are supervised by keypoints, the rest are left for complimentary attention maps: we used 0, 1, 17 and 47 complimentary attention maps.
  \end{tablenotes}
  \label{tab:cub_nattention}
  \end{threeparttable}
  \end{center}
\end{table}

\subsection{Ablation study: Attention Cropping at Test Time}
\label{sec:crop_testtime}

\begin{figure}[t]
    \centering
    \includegraphics[width=0.49\columnwidth,trim={0cm 6cm 0cm 0cm},clip]{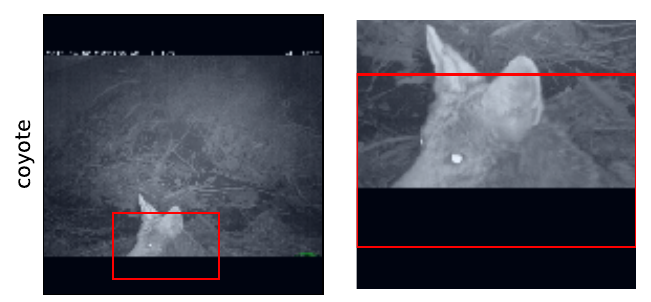}
    \includegraphics[width=0.49\columnwidth,trim={0cm 0cm 0cm 6cm},clip]{figures/CCT_coyote.pdf}
    \caption{Selected Samples for Coyote class, where attention cropping hurts performance. The red bounding boxes are derived from the predicted attention maps.}
    \label{fig:cct_coyote}
\end{figure}

Using the attention maps from \emph{WSDAN} and \emph{PrPool}(ours) it is straight forward to create a bounding box around the areas in the image that the network is attending to. We used this bounding box to crop the image and re-feed it at test time. We observed this is a key element of \emph{WSDAN} and has usually a positive effect, see Table \ref{tab:cub_refeeding}. 
Intuitively this technique should increase the performance as it creates a higher-resolution attention map from the cropped input image (The original image is of size $488^2$ and the attention map is $28^2$).
The results for the CCT20 dataset can be seen in Table \ref{tab:cct_refeeding}, in this case the effect of the re-feeding seems to be hurting the overall performance for both models \emph{WSDAN} and \emph{PrPool}.
We observed class-specific effects. For instance, when using CCT+ as training dataset the rodent class showed an improvement from 20.0\% to 55.0\% after attention cropping at test time (Figure \ref{fig:cct_plus_by_n}). This is likely a similar effect as observed for CUB200, where the high-resolution attention map centered around the (rather small) animal helps the identification; on the other hand, some classes were negatively affected by attention cropping, in particular larger animals such as dog, coyote and raccoon. For these, it sometimes happens that keypoints are not identified correctly and the cropping removes relevant information; see Figure \ref{fig:cct_coyote} for selected samples where this is the case.

\begin{table}[t]
\begin{center}
    \begin{threeparttable}
	\caption{Effect of Re-feeding Attention-cropped Images at Test Time on CUB200 and iBirds{2017} Test Datasets.}
  
        \begin{tabular}{llc*{2}{C{1cm}}}
 Pool & Model   &     Test-crop &  iBirds2017 &  CUB200 \\  \midrule
\multirow{4}{*}{Avg} & \multirow{2}{*}{PrPool} & No &        61.4 &     87.7 \\
    &          & Yes &      \underline{65.2} &     \underline{88.1} \\ \clineh{2-5}
    & \multirow{2}{*}{WSDAN} & No &        54.9 &     87.4 \\
    &          & Yes &        \underline{62.4} &     \underline{88.6} \\\midrule
\multirow{2}{*}{Cov} & \multirow{2}{*}{PrPool} & No &        64.8 &     \underline{\textbf{89.4}} \\
    &          & Yes &       \underline{\textbf{67.0}} &89.0 \\
\end{tabular}

        \begin{tablenotes}
\item Best performance for each case is \underline{underlined}. In \textbf{Bold} the best overall performance. This re-feeding is only possible with \emph{WSDAN} and \emph{PrPool} models.
\end{tablenotes}
	\label{tab:cub_refeeding}
  
\end{threeparttable}
    \end{center}
\end{table}

\begin{table}[t!]
\begin{center}
    \begin{threeparttable}
	\caption{Effect of Re-feeding Attention-cropped Images at Test Time on CCT20 Test datasets}
        \begin{tabular}{llll*{2}{C{0.5cm}}l*{2}{C{0.5cm}}}
& \multirow{2}{*}{Pool}   & \multirow{2}{*}{Model} & \multirow{2}{*}{Test-crop} & \multicolumn{2}{c}{Acc} && \multicolumn{2}{c}{Acc\textsubscript{class}} \\ \cline{5-6} \cline{8-9}
 &   &&  &    CCT+  & CCT  &&     CCT+ & CCT    \\ 
   \multicolumn{7}{l}{\textit{Cis Locations}} \\ \clineh{1-9}
&\multirow{4}{*}{Avg} & \multirow{2}{*}{PrPool} & No &  \underline{\textbf{81.0}} &  \underline{\textbf{83.5}} && \underline{\textbf{69.4}} & 70.3  \\
 &   &        & Yes  &      80.3 &  82.1 &&      66.8 & \underline{\textbf{73.3}}  \\ \clineh{3-9}
  &  & \multirow{2}{*}{WSDAN} & No &      \underline{72.9} &  \underline{76.2} &&      \underline{52.2} &  63.9  \\
   & &        & Yes  &      71.7 &  75.7 &&      50.8 & \underline{ 64.9}  \\ \clineh{2-9}
&\multirow{2}{*}{Cov} & \multirow{2}{*}{PrPool} & No &     \underline{ 76.7 }&  \underline{82.9} &&     \underline{ 57.0} & \underline{ 69.9}  \\
&    &        & Yes  &      68.8 &  75.9 &&      46.7 &  56.3 \\

   \multicolumn{7}{l}{\textit{Trans Locations}} \\ \clineh{1-9} 
&\multirow{4}{*}{Avg} & \multirow{2}{*}{PrPool} & No &           70.6 &  \underline{72.3} &&     52.7 &  54.8 \\
&    &        & Yes  &          \underline{\textbf{72.0}} &  68.5 &&     \underline{\textbf{55.1}} &  \underline{\textbf{60.2}} \\ \clineh{3-9}
 &   & \multirow{2}{*}{WSDAN} & No &        \underline{62.6} &  61.4 &&     \underline{34.7} &  42.4 \\
  &  &        & Yes  &           60.3 & \underline{ 62.5} &&      33.5 &  \underline{42.5} \\ \clineh{2-9} 
&\multirow{2}{*}{Cov} & \multirow{2}{*}{PrPool} & No &        \underline{68.5} &  \underline{\textbf{75.0}} &&      \underline{42.8} & \underline{ 51.6} \\
&    &        & Yes  &            64.8 &  70.5 &&     41.2 &  49.5 \\

\end{tabular}

\begin{tablenotes}
\item  CCT dataset has 13K samples, including 1K samples with keypoints annotations. CCT+ only has 1K, all of them with keypoints annotations. Best performance for each case is \underline{underlined}. In \textbf{Bold} the best overall performance. This re-feeding is only possible with \emph{WSDAN} and \emph{PrPool} models.
\end{tablenotes}
	\label{tab:cct_refeeding}

\end{threeparttable}
\end{center}

\end{table}

\subsection{Qualitative Evaluation of Attention Maps}

\begin{figure}[t]
    \centering
    \includegraphics[width=\columnwidth]{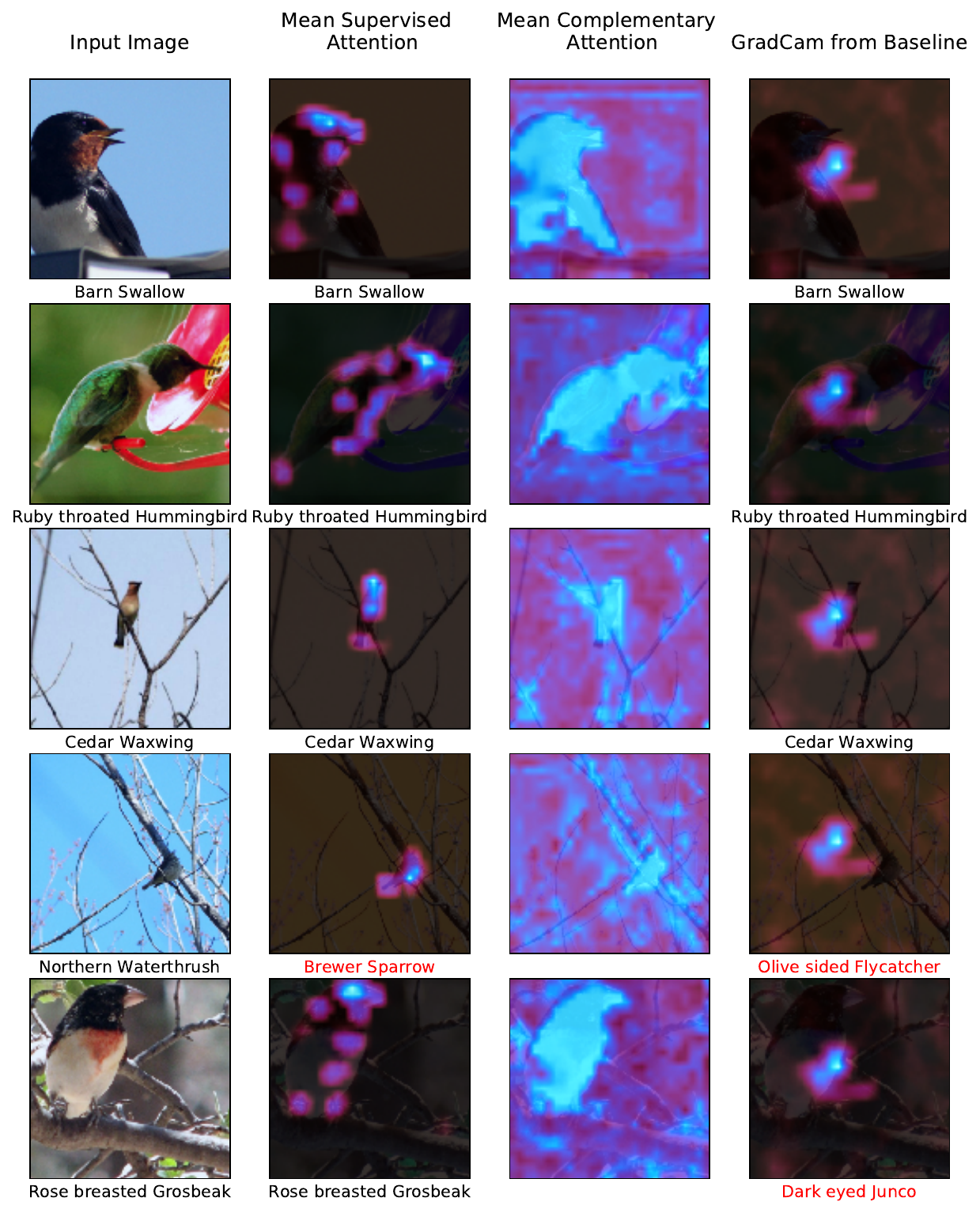}
    \caption{Comparison GradCam from \emph{AvgPool} and Mean Attention Maps from \emph{CovPrPool}. Random samples from iBirds{2017} test dataset. Left most column is input image. Below each sample GT class, and predictions are shown. Misclassified samples in \textcolor{red}{red}}
    \label{fig:gradcam_atn}
\end{figure}

Attention Maps make it easy to visualize which parts of the input image are being used to make a prediction. 
For \emph{CovPrPool}, we computed the mean of the attention maps (separately for the supervised and the complementary maps). For the baseline model \emph{AvgPool} with ResNet101 backbone, we used GradCam \cite{selvaraju2017grad}. In Figure~\ref{fig:gradcam_atn}, we show random examples from the iBirds{2017} dataset. One can see that the baseline focusses on small, specific patterns on the bird, in some cases even on areas not on the bird (see fourth row from Figure~\ref{fig:gradcam_atn}). This provides an intuitive example how the privileged information can help generalization, by paying attention to relevant, representative parts of the bird.

Furthermore, we can see that the Complementary attention map learn to largely ignore pixels outside the bird's silhouette, despite not being explicitly trained for this.

\begin{figure}[t]
    \centering
    \includegraphics[width=0.9\columnwidth]{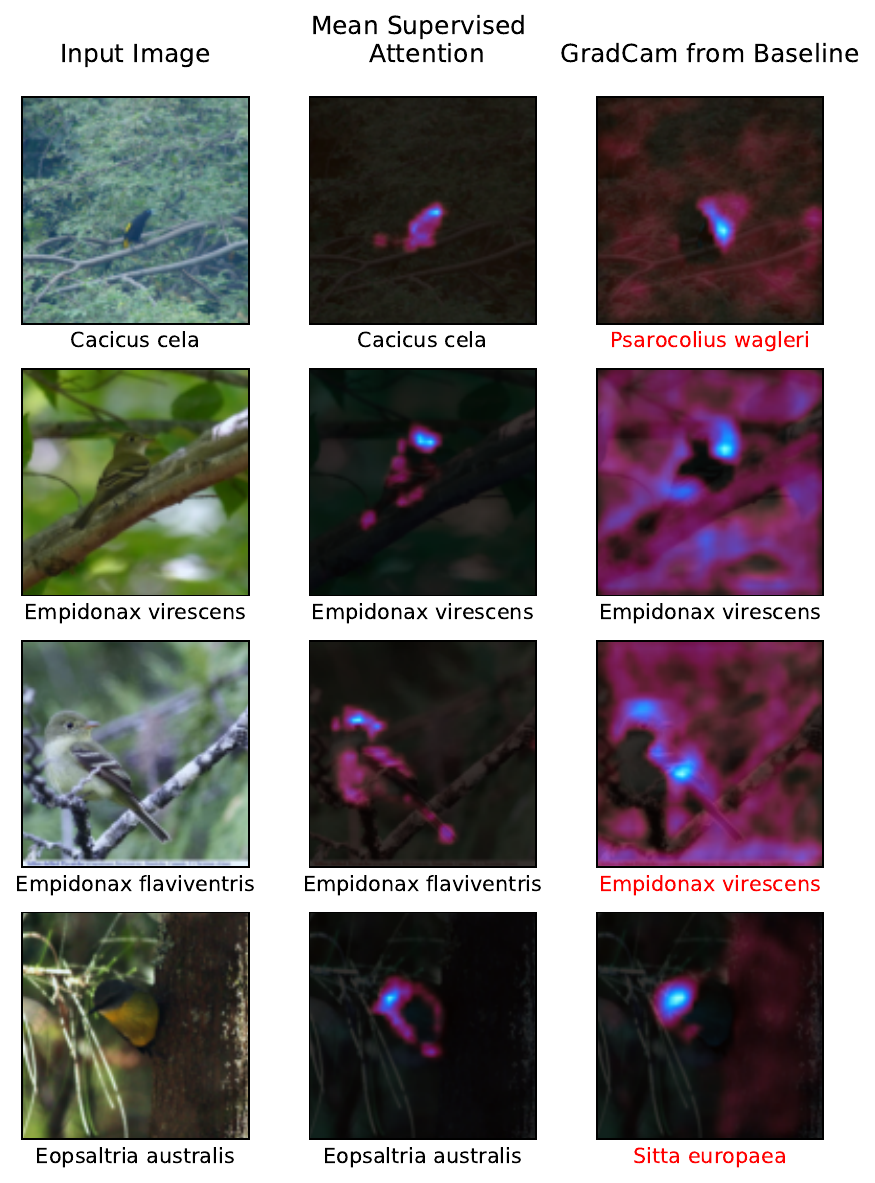}
    \caption{{Comparison GradCam from \emph{AvgPool} and Mean Attention Maps from \emph{CovPrPool}. Selected samples from low-shot classes from iBirds{2018} test dataset. Left most column is input image. Below each sample GT class, and predictions are shown. Misclassified samples in \textcolor{red}{red}}}
    \label{fig:gradcam_atn_ibirds18}
\end{figure}

{We show selected test samples from low-shot of iBirds2018 in Figure~\ref{fig:gradcam_atn_ibirds18}. The attention maps with \emph{AvgPool} show slight overfitting to background features, which \emph{CovPrPool} seems to effectively ignore.}

\begin{figure*}[t]
    \centering
    	\caption{Random samples from CCT Cis Test Set. Red bounding-box is derived from the predicted attention maps. From left to right: Input image ({marked }with predicted class), zoom to attended region and keypoint specific attention maps.}
    \label{fig:cct_random_samples}
    \includegraphics[width=0.9\textwidth]{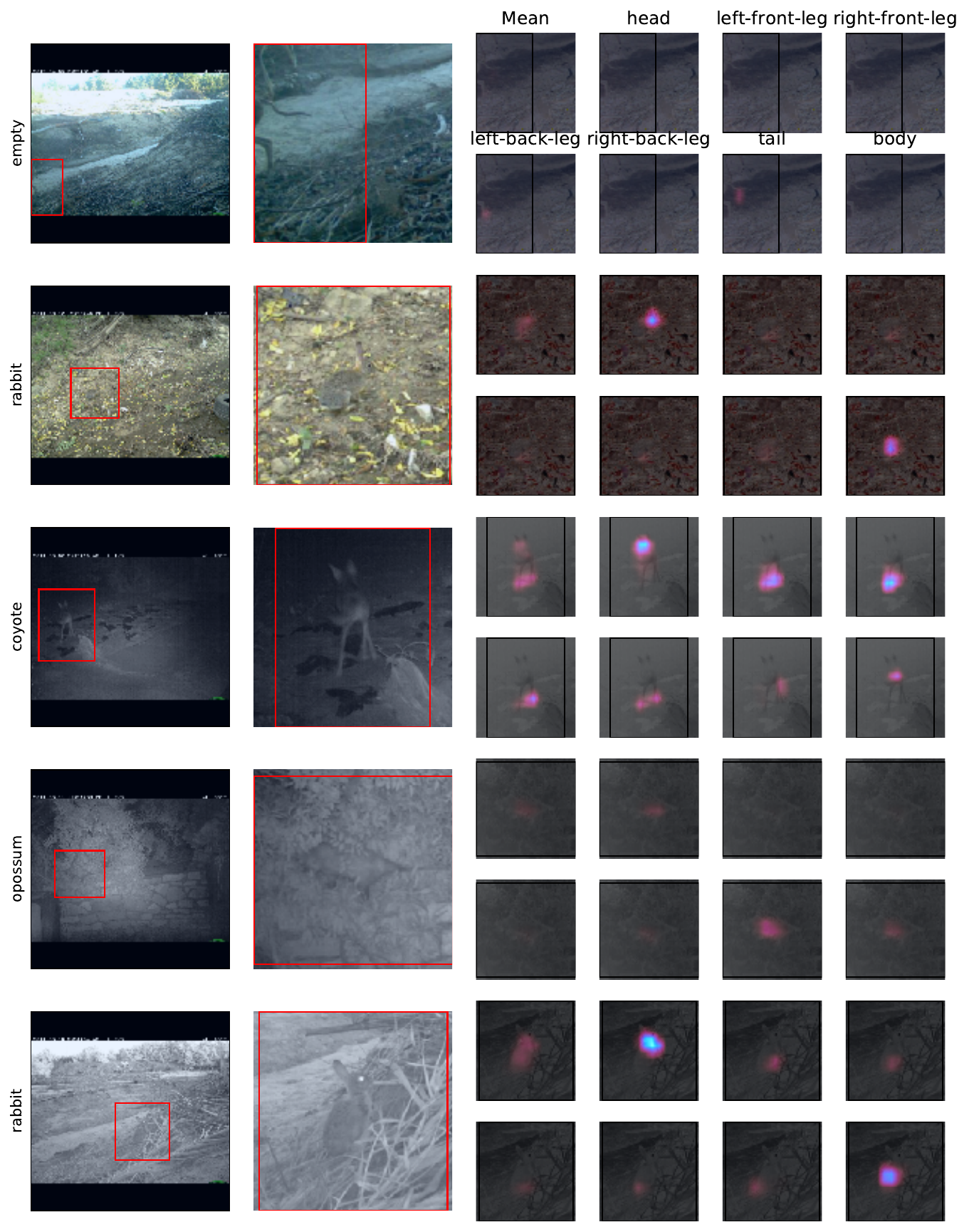}
\end{figure*}

\begin{figure*}[t]
    \caption{Random samples from iBirds{2017} Test Set. The red bounding-box is derived from the predicted attention maps. From left to right: Input image (with predicted class and misclassified samples {include true class in parenthesis}), zoom to attended region, keypoint specific attention maps, {and average of complementary attention maps.}}
    \label{fig:ibirds_random_samples}
    \centering
    \includegraphics[width=\textwidth]{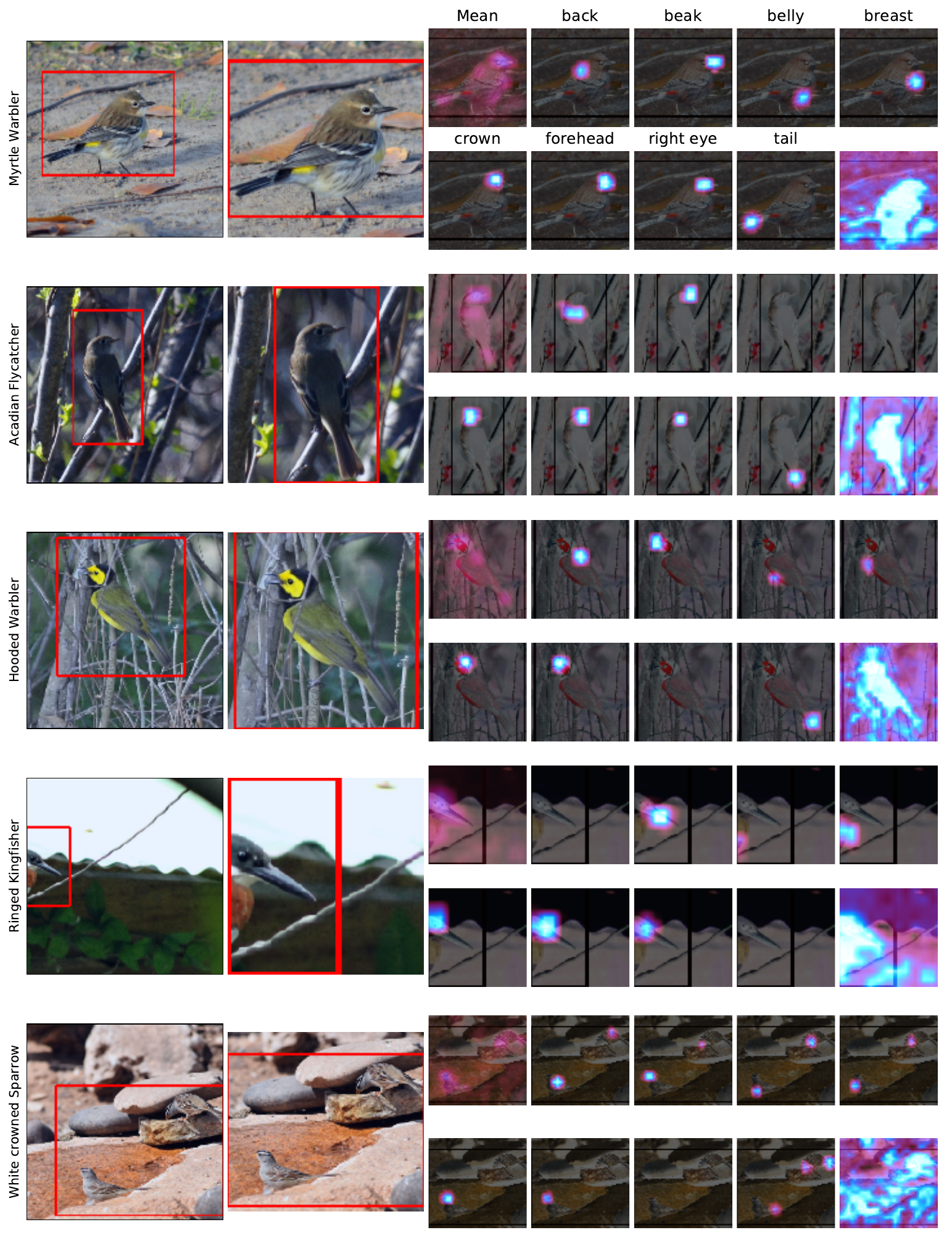}
\end{figure*}

\textbf{Example attention maps of trained models}.
See Figure~\ref{fig:cct_random_samples} for random samples of predictions and attention maps over the CCT Cis Test set. For these samples, we observe that the attention maps clearly highlight the different keypoints, even when the animal is difficult to distinguish from the background. Figure~\ref{fig:ibirds_random_samples}, on the other hand, shows samples from the iBirds{2017} test dataset. Here the attention map on the bottom right is complementary to the keypoints and effectively performs a foreground-background separation.

\section{Conclusions}

The aim of learning under privileged information is to exploit collateral information that is available only for the training data, so as to learn predictors that generalize better.
We have examined the case where the privileged information comes in the form of keypoint locations, a natural and fairly frequent situation in image analysis.
By using keypoints as supervision for attention maps, they can be effectively leveraged to support image classification.
Privileged information to steer a model's attention is particularly effective when labeled training data is scarce, and when it exhibits strong biases.
Moreover, it turns out that in some small-data scenarios a moderate amount of privileged information may serve as an alternative to few-shot learning.

On a more general note, we see it as an important message of our work that gathering more data is not the only option to fix an under-trained deep learning model. While additional training data is almost always welcome, there are important applications where it is inherently hard to come by.
It is encouraging that, with the right design, more elaborate labeling of the existing data can also present a way forward.

\ifCLASSOPTIONcompsoc
\else
\fi


\ifCLASSOPTIONcaptionsoff
  \newpage
\fi



\bibliographystyle{IEEEtran}

\bibliography{egbib}
%
%
%

%

\begin{IEEEbiography}[{\includegraphics[width=1in,height=1.25in,clip,keepaspectratio]{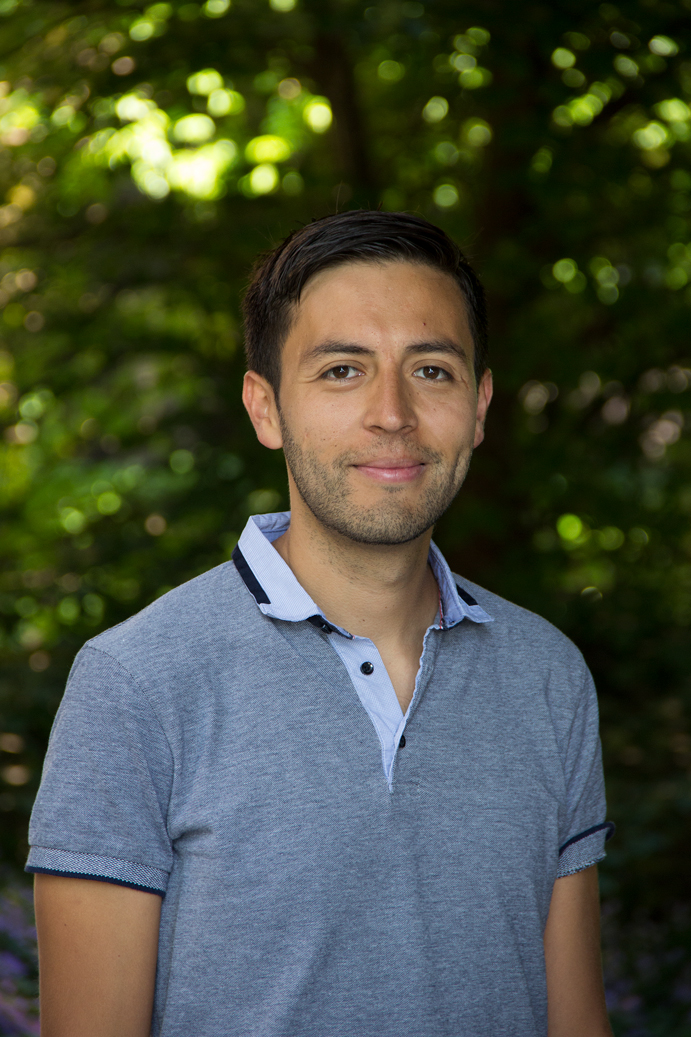}}]{Andr\'es C Rodr\'iguez}
 recieved his BS in Industrial Engineering from the Pontificia Universidad Javeriana in Bogota, Colombia in 2012. Then recieved his MSc in Statistics from ETH Zurich in 2017. He joined the EcoVision Lab at ETH Zurich in 2017 as a PhD student.
	 His research interest is at the intersection of machine learning and remote sensing with the focus on large scale analysis for biodiversity, sustainablity and environmental applications.
\end{IEEEbiography}

%
%
%


\begin{IEEEbiography}[{\includegraphics[width=1in,height=1.25in,clip,keepaspectratio]{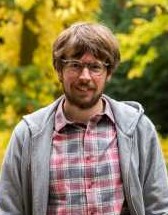}}]{Stefano D'Aronco}
received his BS and MS degrees in electronic engineering from the Università degli studi di Udine, in 2010 and 2013 respectively. He then joined the Signal Processing Laboratory (LTS4) in 2014 as a PhD student under the supervision of Prof. Pascal Frossard. He received his PhD in Electrical Engineering from École Polytechnique Fédérale de Lausanne in 2018. He is Postdoctoral researcher in the EcoVision group at ETH Zurich since 2018. His research interests include several machine learning topics, such as Bayesian inference method and deep learning, with particular emphasis on applications related to remote sensing an environmental monitoring.
\end{IEEEbiography}

\begin{IEEEbiography}[{\includegraphics[width=1in,height=1.25in,clip,keepaspectratio]{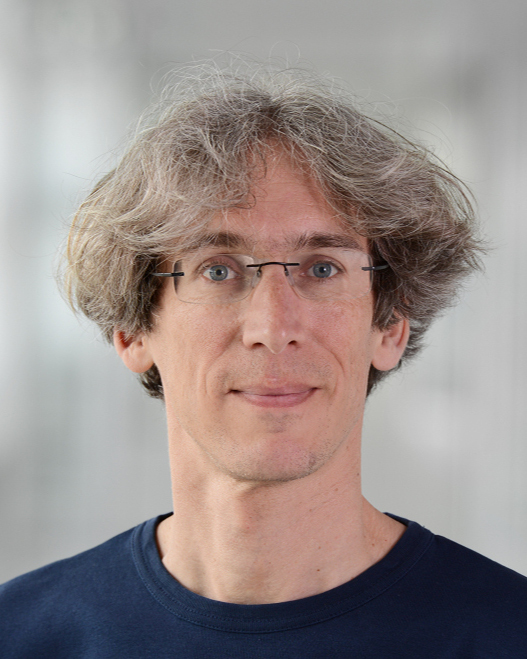}}]{Konrad Schindler}
(M'05–SM'12) received the Diplomingenieur (M.Tech.) degree from Vienna University of Technology, Vienna, Austria, in 1999, and the Ph.D.\ degree from Graz University of Technology, Graz, Austria, in 2003.
He was a Photogrammetric Engineer in the private industry and held researcher positions at Graz University of Technology, Monash University, Melbourne, VIC, Australia, and ETH Zu\"urich, Z\"urich, Switzerland. He was an Assistant Professor of Image Understanding with TU Darmstadt, Darmstadt, Germany, in 2009. Since 2010, he has been a Tenured Professor of Photogrammetry and Remote Sensing with ETH Z\"urich. His research interests include computer vision, photogrammetry, and remote sensing.
\end{IEEEbiography}

\begin{IEEEbiography}[{\includegraphics[width=1in,height=1.25in,clip,keepaspectratio]{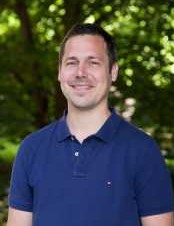}}]{Jan Dirk Wegner}
 is associate professor at University of Zurich and head of the EcoVision Lab at ETH Zurich. Jan was PostDoc (2012-2016) and senior scientist (2017-2020) in the Photogrammetry and Remote Sensing group at ETH Zurich after completing his PhD (with distinction) at Leibniz Universität Hannover in 2011. He was granted multiple awards, among others an ETH Postdoctoral fellowship and the science award of the German Geodetic Commission. Jan was selected for the WEF Young Scientist Class 2020 as one of the 25 best researchers world-wide under the age of 40 committed to integrating scientific knowledge into society for the public good. He is vice-president of ISPRS Technical Commission II, chair of ISPRS II/WG 6 "Large-scale machine learning for geospatial data analysis", director of the PhD graduate school "Data Science" at University of Zurich, associated faculty of the ETH AI Center, and his professorship is part of the Digital Society Initiative at University of Zurich.
\end{IEEEbiography}

\vfill

\clearpage 
\appendices

\section{Inception vs ResNet Backbone}
\label{sec:wsdan_backbone}

\begin{table}[h]
\begin{center}
    \begin{threeparttable}
	\caption{Top1-Accuracy of WSDAN with ResNet-101 and PrPool (ours) on CUB200 and iBirds{2017} Test Datasets.}
        \begin{tabular}{lll*{2}{C{1cm}}}
Pool & Model & Backbone   &  CUB200 &  iBirds2017 \\
\midrule
\multirow{1}{*}{Avg}  & PrPool & ResNet-101 &    87.7 &    64.3 \\
    & WSDAN & InceptionV3 &    88.0 &    60.4 \\
    & WSDAN &ResNet-101 &    85.8 &    55.1  \\ \midrule
\multirow{1}{*}{Cov}  & PrPool & ResNet-101 &    89.2 &    67.9 \\
\end{tabular}

	\label{tab:wsdan_backbone}
\end{threeparttable}
    \end{center}
\end{table}

\section{{Complementary Loss}}
\label{sec:compl_loss}

\begin{figure*}[t]
    \centering
    \includegraphics[width=\textwidth]{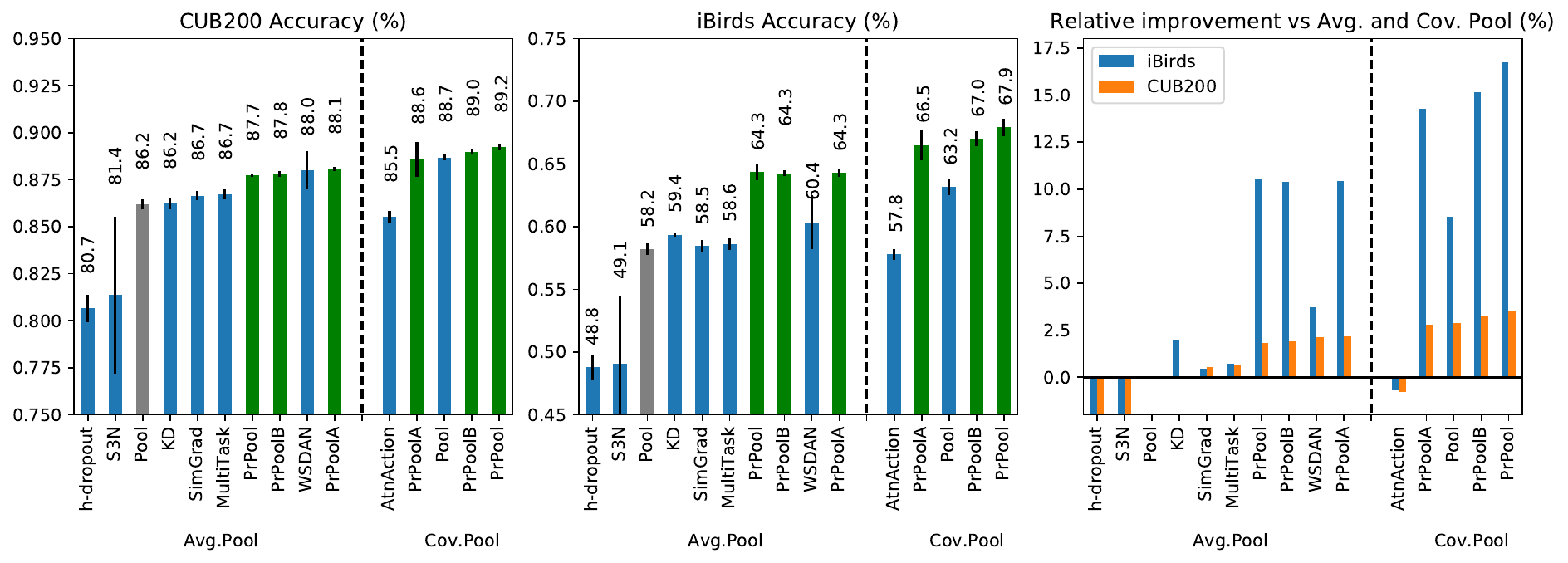}
    \caption{ {Top-1 accuracy for CUB (left) and iBirds{2017} (middle) test datasets. {Mean accuracy and standard deviation (error bars) over 5 runs.} In gray the baselines methods \emph{AvgPool} and \emph{CovPool}; green indicates our methods trained with \emph{PrPool} (ours). 
    Right: Relative improvement vs the baseline method Avg.Pool. PrPoolA and PrPoolB represent different implementations of the loss $l_{reg}(\v{a}_q)$}}
    \label{fig:cub_full_losses}
\end{figure*}

{
In Figure~\ref{fig:cub_full_losses} we have added models trained with variations of $l_{reg}(a_q)$. \emph{PrPool} were trained using the definition described in the methods section. \emph{PrPoolA} and \emph{PrPoolB}, were trained using Loss A and Loss B, respectively defined as follows:}

\begin{itemize}
\item
Loss A ``Overlap penalization": 
\begin{equation*}
l_{reg}(\v{a}_q) = \sum_{q': q\neq q'}^Q  \min(\Delta-\| \Gamma_{q} - \Gamma_{q'}\|,0)^2,
\end{equation*}
where $\Gamma_{q}$ denotes the center of mass of the attention map $q$. Note that $\| \Gamma_{q} - \Gamma_{q'}\|$ is a non-negative amount.
The center of mass  $\Gamma_{q}$ can be computed as: 

\begin{equation*}
\Gamma_{q} = \frac{1}{WH} \sum_{WH} \text{Grid}_{wh}a_{whq},
\end{equation*}
where $ \text{Grid}_{wh}$ correspond to a coordinate grid map.
\vspace{1em}
\item Loss B "Variance revised":
\begin{equation*}
l_{reg}(\v{a}_q) = \frac{1}{WH}\sum_{WH}(\v{a}_{whq} - \bar{\v{a}}_q)^2
\end{equation*}

\end{itemize}

{
Loss A aims to penalize \emph{attention centers} $\Gamma_{q}$ that lie too close to each other. We compute all centers of attention $\Gamma_{q}$ and penalize any two attention maps that lie within a distance $\Delta$. Loss B is an alternative formulation that encourages variance inside each attention map. We present the results in Figure~\ref{fig:cub_full_losses}, with PrPool denoting the basic loss proposed in eq.~(\ref{eq:reg_loss}). Although both alternative formulations A and B encourage complementarity of attention maps, which is in principle desirable, we do not observe significant improvements over the simpler $l_{reg}(\v{a}_q)$; and all three loss variants outperform the baselines on the iBirds{2017} dataset.
}

\begin{table}[t]
\begin{center}
    \begin{threeparttable}
	\caption{{ImageNet subsets. Bird subset, and classes on CUB200.}}
	    \label{tab:imagenet_cub}
\begin{tabular}{llrrr}
     & \multirow{2}{*}{}  & \multirow{2}{*}{\#classes} & \multicolumn{1}{c}{avg \#imgs} & \multicolumn{1}{c}{total \#imgs} \\
     &  &   & \multicolumn{1}{c}{per class}  & \multicolumn{1}{c}{($\times$1000)}  \\
    \midrule
    \multicolumn{3}{l}{\itshape ImageNet21k} \\\clineh{1-5}
     & ---  &  21841  & 650.0  &  14195.9 \\
     & Birds &  357  & 994.3  &  355.0\\
     & CUB200 Birds  &  64  & 1108.3  &  70.9\\
    \midrule
    \multicolumn{3}{l}{\itshape ImageNet1k} \\\clineh{1-5}
     & --- &  1000  & 1291.4  &  1290.1\\
     &  Birds &  19  & 1278.4  &  24.3\\
     & CUB200 Birds&  3  & 1244.7  &  3.7\\
\end{tabular}
\begin{tablenotes}
     \item {"Birds" refers to ImageNet-21k classes whose class description includes the scientific name of any bird species, under "CUB200 Birds" that set is further restricted to species contained in CUB200.}
\end{tablenotes}
\end{threeparttable}
    \end{center}
\end{table}

\begin{table}[t]
\begin{center}
    \begin{threeparttable}
	\caption{{Comparison of AvgPrPool, CovPrPool, and TransFG with differently pretrained backbone.}}
\label{tab:TransFG_pretraining}

\begin{tabular}{llcc}
    \multirow{1}{*}{Model}&  Backbone$^*$  & \multirow{1}{*}{CUB200} & \multirow{1}{*}{iBirds2017}  \\
     \midrule
    AvgPool & Resnet-101 (1k) & 86.2  & 58.2 \\
	AvgPrPool & Resnet-101 (1k) & 87.7  & 64.3 \\
    CovPool & Resnet-101 (1k) & 88.7  & 63.2 \\
    CovPrPool & Resnet-101 (1k) & 89.2  & 67.9 \\
    TransFG &  ViTB16 (21k$^\dagger$) &91.3 &  73.4 \\
    TransFG &  ViTB16 (1k$^{\dagger\dagger}$) & 86.1 & 56.6  \\
\end{tabular}
    
\begin{tablenotes}
     \item {*: In parenthesis the size of the pretraining set. $^\dagger$: 21k model from \cite{dosovitskiy2020image}. $^{\dagger\dagger}$: 1k model from \cite{steiner2021train}}
\end{tablenotes}
\end{threeparttable}
    \end{center}
\end{table}

\section{Transformer Baselines}
\label{sec:compl_transformer}

{
\textbf{TransFG\cite{he2021transfg}} reports 91.7\% on CUB200 with a modified transformer network \cite{dosovitskiy2020image}. Transformer networks have weaker inductive biases than CNNs, and come into their own when trained on much larger datasets. Typically, they are pretrained on ImageNet-21k, as opposed to CNNs that are usually pretrained on a (balanced) subset with only 1k classes.}

{However, a larger training set also means a bigger risk of inadvertedly exposing information that is directly relevant for a specific test set. Our focus is on the downstream task of fine-grained bird classification, where the standard benchmark has, for some time, been CUB200. In Table~\ref{tab:imagenet_cub} we compare the two variants of ImageNet with CUB200, in terms of the number of classes, the number of images, and the class overlap.
It turns out that ImageNet-21K contains a vastly larger number of birds than its 1k subset, in total across all species \textgreater350k samples. Among those, \textgreater70k samples are from bird species contained in CUB200. Note that this is an order of magnitude more than the entire training set of CUB200 itself (5994 samples).}

{
This large degree of overlap raises the question whether it is at all reasonable to employ models pretrained on ImageNet-21k when doing experiments on CUB200. Obviously, any comparison between a model that has been pretrained om ImageNet-21k and another one that has not will be significantly skewed. In particular investigations about few-shot learning appear meaningless with a model that has, during its pretraining, seen more examples of birds than what the complete CUB200 training set can provide. To complement the dataset comparison, Table~\ref{tab:TransFG_pretraining} compares TransFG with our proposed method. Our experiment confirms the excellent classification performance of TransFG also on iBirds{2017}, but it is difficult to ascertain to what degree that result is due to its pretraining. Interestingly, when pretrained only on ImageNet-1k, TransFG performs pretty much on par with the Avg.Pool baseline. Lastly, we tested TransFG (pretrained on the full ImageNet-21k) also on CCT, and observed poor performance on both the Cis and the Trans test set, see Table~\ref{tab:CCT_w_transfg}}.

\begin{table}[t]
    \centering
    \caption{Overall Accuracy and Mean per Class Accuracy Results of Models Trained on CCT20 and CCT20+. TransFG baseline in blue}
    \label{tab:CCT_w_transfg}

\begin{tabular}{lllC{0.5cm}C{0.5cm}lC{0.5cm}C{0.5cm}}
&    & {} & \multicolumn{2}{c}{Acc} && \multicolumn{2}{c}{Acc\textsubscript{class}}  \\ \cline{4-5} \cline{7-8}
 &   & Dataset &    CCT+  & CCT && CCT+ & CCT    \\
\midrule
     \multicolumn{8}{l}{\textit{Cis Locations}} \\ \clineh{1-8}
&\multirow{5}{*}{Avg} 
    & S3N &      75.2 &  75.5 &&     61.7 &  66.7   \\
    &   & Pool &      71.4 &  73.6 &&      57.0 &  59.8         \\
   & & MultiTask &      72.2 &  76.4 &&      59.5 &  65.8         \\
 && WSDAN &      72.9 &  76.2 &&      52.2 &  63.9   \\
    &  & PrPool  (ours) & \textbf{81.0} & 	\textbf{83.5}&& 	\textbf{69.4}& 	\textbf{70.3} \\
       \clineh{2-8}
&\multirow{2}{*}{Cov} & Pool &      74.3 &  81.4 &&      59.8 &  69.3   \\
 &   & AtnAction &    72.7 & 	77.4 && 	56.3 & 	68.0   \\
 &   & PrPool (ours) &      76.7 &	82.9&& 	57.0& 	69.9  \\\clineh{2-8}
 & ViT   & \textcolor{blue}{TransFG(21k)} &   69.9 &	76.3 && 	59.8 & 	66.8  \\
 \multicolumn{8}{l}{\textit{Trans Locations}} \\ \clineh{1-8} 
&\multirow{5}{*}{Avg} 
 & S3N &         64.2 &  70.6 &&      40.2 &  50.9 \\
&& Pool &           65.1 &  66.1 &&      42.0 &  43.6 \\
&& MultiTask &           65.6 &  67.2 &&      42.6 &  50.0 \\
&& WSDAN &        62.6 &  61.4 &&      34.7 &  42.4 \\
&& PrPool (ours) &  \textbf{70.6} & 	72.3 && 	\textbf{52.7} &  \textbf{54.8} \\
     \clineh{2-8}
&\multirow{2}{*}{Cov} & Pool &            67.2 &  70.1 &&      41.9 &  43.8 \\
 &   & AtnAction &    64.4 & 	73.0 && 	43.2 & 	51.2  \\
 &   & PrPool (ours) &       68.5 & 	\textbf{75.0}&& 	42.8& 	51.6 \\\cline{2-8} \rule{0pt}{3ex}
  & ViT  & \textcolor{blue}{TransFG(21k)} &   62.8 &	66.5 &&  40.78   & 	48.6  \\
\end{tabular}
\end{table}

\section{Few-shot baselines}
\label{sec:compl_fewshot}

{See Table~\ref{tab:additional_fewshot} for additional baselines regarding few-shot learning subsets in CUB200 and iBirds{2017}.}

\begin{table}[t]
\begin{center}
    \begin{threeparttable}
	\caption{Additional baselines for {CUB200 (top) and iBirds{2017} (bottom). The values are top-1 accuracies with 100 base classes and 100 new classes with $n$-shots.}}
        \begin{tabular}{lL{1.3cm}L{1.7cm}*{4}{C{0.5cm}}}
 & \multirow{2}{*}{Pool}   & \multirow{2}{*}{Model} & \multicolumn{4}{c}{N-shot} \\\cline{4-7}
& &  &    1  &   2  &   5  &   10 \\
\midrule
\multicolumn{6}{l}{\textit{CUB200}} \\\clineh{1-7}
&Avg & MultiTask &    48.1 & 56.7 & 73.7 & 81.6 \\
 &     & Pool &    48.3 & 56.6 & 73.4 & 81.4 \\
  &    & PrPool &    52.4 & \underline{64.5} & \textbf{\underline{78.8}} & 84.5 \\
   &   & S3N &    50.4 & 60.2 & 76.3 & 82.7 \\
    &  & SimGrad &    48.3 & 56.8 & 73.8 & 81.7 \\
     & & WSDAN &    \underline{52.7} & 63.7 & \textbf{\underline{78.8}} & \textbf{85.0} \\\clineh{2-7}
&Cov & AtnAction &    49.8 & 60.5 & 75.3 & 81.5 \\
 &     & Pool &    48.9 & 58.5 & 76.2 & 84.2 \\
  &    & PrPool &    49.1 & 59.7 & 77.9 & \underline{84.9} \\\clineh{2-7} 
&Other & FewShot &    \textbf{63.1} & \textbf{69.2} & 74.5 & 76.9 \\
\midrule
\multicolumn{6}{l}{\textit{iBirds2017}} \\\clineh{1-7}
&Avg & MultiTask &       33.9 & 37.5 & 47.1 & 53.3 \\
 &     & Pool &       34.2 & 37.7 & 47.1 & 53.7 \\
  &    & PrPool &       \textbf{39.1} & \textbf{46.4} & \underline{54.8} & \underline{59.9}\\
   &   & S3N &       34.6 & 39.7 & 48.3 & 54.2 \\
    && SimGrad &       34.2 & 38.2 & 47.6 & 53.5 \\
    &  & WSDAN &       37.6 & \underline{43.0} & 52.5 & 57.7 \\\clineh{2-7}
&Cov & AtnAction &       34.3 & 39.7 & 48.1 & 52.9 \\
 &     & Pool &       35.8 & 40.2 & 50.9 & 57.5 \\
  &    & PrPool &       \underline{37.8} & \underline{43.0} & \textbf{55.3} & \textbf{61.8} \\
\end{tabular}

       \begin{tablenotes}
       \item {Average over five random novel/base splits.
       \textbf{Bold} and \underline{underlined} numbers denote the best and second best overall performance. FewShot results are taken directly from~\cite{tokmakov2018learning}}
 \end{tablenotes}
	\label{tab:additional_fewshot}
\end{threeparttable}
    \end{center}
\end{table}


\end{document}